\documentclass[11pt,a4paper]{article}
\usepackage[hyperref]{naaclhlt2019}
\usepackage{times}
\usepackage{latexsym}
\usepackage{url}

\usepackage{graphicx}
\usepackage{booktabs}
\usepackage{multirow}
\usepackage{cleveref}
\usepackage{xcolor}
\usepackage{todonotes}
\usepackage[toc,page]{appendix}
\usepackage{xspace}
\usepackage{comment}
\usepackage{linguex}
\usepackage{enumitem}

\newcommand{\appref}[1]{Appendix~\ref{sec:#1}}
\newcommand{\labeledsection}[2]{\section{#1}\label{sec:#2}}
\newcommand{\cwr}[0]{\textsc{cwr}\xspace}
\newcommand{\cwrs}[0]{\textsc{cwr}s\xspace}

\aclfinalcopy %
\def\aclpaperid{2049} %

\let\svthefootnote\thefootnote
\newcommand\blankfootnote[1]{%
  \let\thefootnote\relax\footnotetext{#1}%
  \let\thefootnote\svthefootnote%
}

\title{Linguistic Knowledge and Transferability of Contextual Representations}

\author{ 
    Nelson F. Liu$^{\spadesuit\heartsuit}$\footnotemark[1] \quad
	Matt Gardner$^{\clubsuit}$ \quad
	Yonatan Belinkov$^{\diamondsuit}$ \\
	{\bf Matthew E. Peters}$^{\clubsuit}$ \quad
	{\bf Noah A.~Smith}$^{\spadesuit\clubsuit}$ \\
    $^\spadesuit$Paul G. Allen School of Computer Science \& Engineering,\\ University of Washington, Seattle, WA, USA \\
	$^\heartsuit$Department of Linguistics, University of Washington, Seattle, WA, USA \\
	$^\clubsuit$Allen Institute for Artificial Intelligence, Seattle, WA, USA \\
	$^\diamondsuit$Harvard John A.\ Paulson School of Engineering and Applied Sciences and\\ MIT Computer Science and Artificial Intelligence Laboratory, Cambridge, MA, USA \\
{\tt \{nfliu,nasmith\}@cs.washington.edu} \\
{\tt \{mattg,matthewp\}@allenai.org}, \hspace{1em} 
{\tt belinkov@seas.harvard.edu}
}

\date{}

\begin{document}
\maketitle
\blankfootnote{\llap{\textsuperscript{*}}Work done while at the Allen Institute for Artificial Intelligence.}
\begin{abstract}
Contextual word representations derived from large-scale neural language models are successful across a diverse set of NLP tasks, suggesting that they encode useful and transferable features of language.
To shed light on the linguistic knowledge they capture, we study the representations produced by several recent pretrained contextualizers (variants of ELMo, the OpenAI transformer language model, and BERT) with a suite of seventeen diverse probing tasks.
We find that linear models trained on top of frozen contextual representations are competitive with state-of-the-art task-specific models in many cases, but fail on tasks requiring fine-grained linguistic knowledge (e.g., conjunct identification).
To investigate the transferability of contextual word representations, we quantify differences in the transferability of individual layers within contextualizers, especially between recurrent neural networks (RNNs) and transformers. For instance, higher layers of RNNs are more task-specific, while transformer layers do not exhibit the same monotonic trend.
In addition, to better understand what makes contextual word representations transferable, we compare language model pretraining with eleven supervised pretraining tasks.  For any given task, pretraining on a closely related task yields better performance than language model pretraining (which is better on average) when the pretraining dataset is fixed. However, language model pretraining on \emph{more data} gives the best results.
\end{abstract}

\section{Introduction}

\begin{figure}[t]
	\includegraphics[width=0.95\columnwidth]{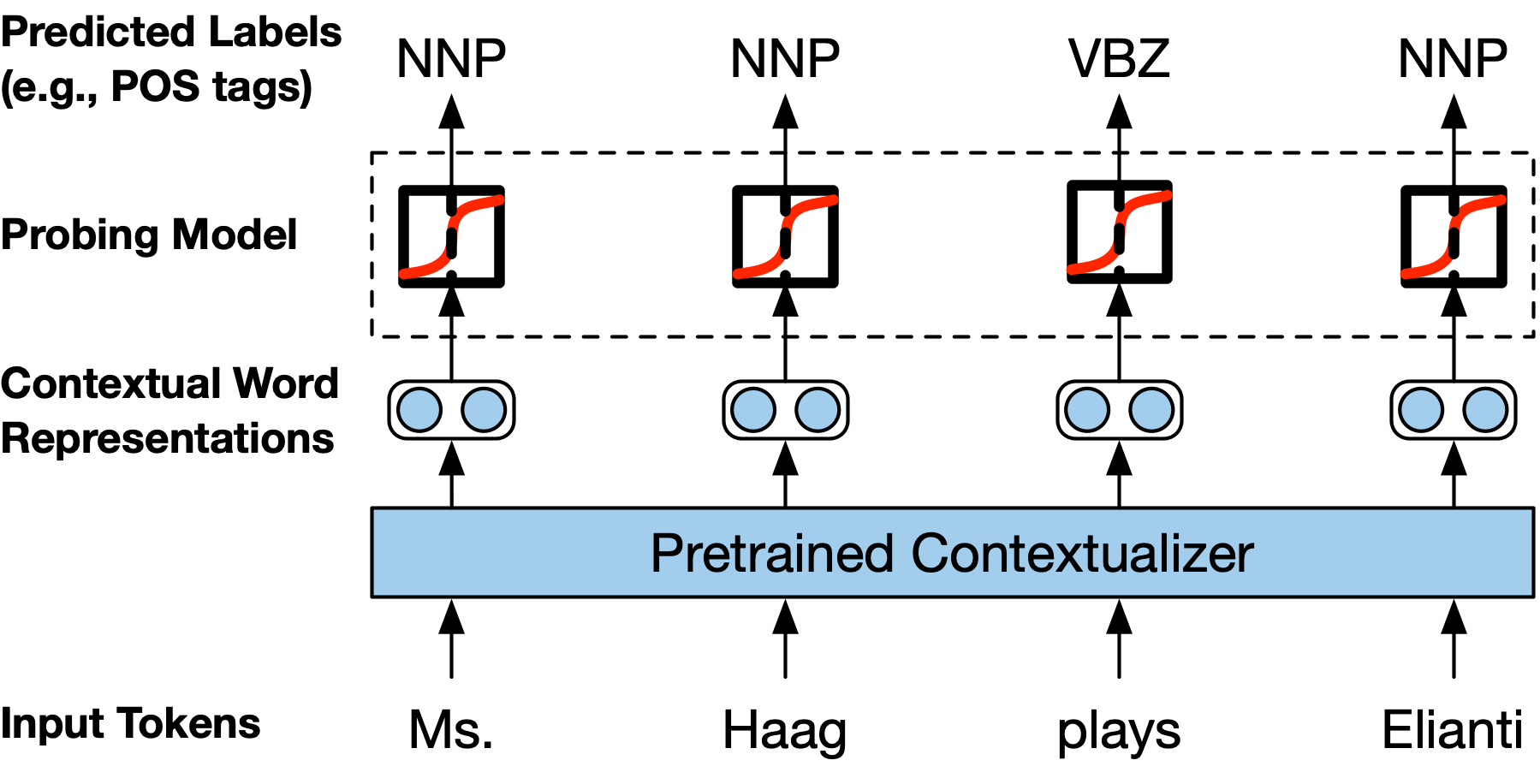}
	\caption{An illustration of the probing model setup used to study the linguistic knowledge within contextual word representations.}
\label{fig:probing_setup}
\end{figure}

Pretrained word representations \citep{Mikolov2013DistributedRO,pennington2014glove} are a key component of state-of-the-art neural NLP models.
Traditionally, these word vectors are static---a single vector is assigned to each word.
Recent work has explored \emph{contextual} word representations (henceforth: \cwrs), which assign each word a vector that is a function of the entire input sequence; this enables them to model the use of words in context.
\cwrs are typically the outputs of a neural network  (which we call a \emph{contextualizer}) trained on tasks with large datasets, such as machine translation \citep{mccann2017learned} and language modeling \citep{Peters2018DeepCW}.
\cwrs are extraordinarily effective---using them in place of traditional static word vectors within the latest models leads to large gains across a variety of NLP tasks.

The broad success of \cwrs indicates that they encode useful, transferable features of language. However, their \emph{linguistic knowledge} and \emph{transferability} are not yet well understood.

Recent work has explored the linguistic knowledge captured by language models and neural machine translation systems,
but these studies often focus on a single phenomenon, e.g., knowledge of hierarchical syntax \citep{blevins2018deep} or morphology \citep{Belinkov2017WhatDN}. We extend prior work by studying \cwrs with a diverse set of seventeen probing tasks designed to assess a wide array of phenomena, such as  coreference, knowledge of semantic relations, and entity information, among others. The result is a broader view of the linguistic knowledge encoded within \cwrs.

With respect to transferability, pretraining contextualizers on the language modeling task has had the most empirical success, but we can also consider pretraining contextualizers with other supervised objectives and probing their linguistic knowledge. We examine how the  pretraining task affects the linguistic knowledge learned,  considering twelve pretraining tasks and assessing transferability to nine target tasks.

Better understanding the linguistic knowledge and transferability of \cwrs is necessary for their principled enhancement through new encoder architectures and pretraining tasks that build upon their strengths or alleviate their weaknesses \citep{linzen2018can}.
This paper asks and answers:

\begin{enumerate}[itemsep=0.5mm]
\item What features of language do these vectors capture, and what do they miss? (\S\ref{sec:linguistic_knowledge})
\item How and why does transferability vary across representation layers in contextualizers? (\S\ref{sec:layerwise_transferability})
\item How does the choice of pretraining task affect the vectors' learned linguistic knowledge and transferability? (\S\ref{sec:pretraining_task_transferability})
\end{enumerate}

We use probing models\footnote{Sometimes called auxiliary or  diagnostic classifiers.} \citep{shi2016does,adi2016fine,Hupkes2018VisualisationA,belinkov2019analysis} to analyze the linguistic information within \cwrs. Concretely, we generate features for words from pretrained contextualizers and train a model to make predictions from those  features alone (\Cref{fig:probing_setup}). If a simple model can be trained to predict linguistic information about a word (e.g., its part-of-speech tag) or a pair of words (e.g., their semantic relation) from the \textsc{cwr}(s) alone, we can reasonably conclude that the \textsc{cwr}(s) encode this information.

Our analysis reveals interesting insights such as: %
\begin{enumerate}[itemsep=0.5mm]
    \item Linear models trained on top of frozen \cwrs are competitive with state-of-the-art task-specific models in many cases, but fail on tasks requiring fine-grained linguistic knowledge. In these cases, we show that \emph{task-trained contextual features} greatly help with encoding the requisite knowledge.
    \item The first layer output of long short-term memory (LSTM) recurrent neural networks is consistently the most transferable, whereas it is the middle layers for transformers.
    \item Higher layers in LSTMs are more task-specific (and thus less general), while the transformer layers do not exhibit this same monotonic increase in task-specificity. 
    \item Language model pretraining yields representations that are more transferable \emph{in general} than eleven other candidate pretraining tasks, though pretraining on related tasks yields the strongest results for individual end tasks.
\end{enumerate}

\section{Probing Tasks}
\label{sec:probing_tasks}

We construct a suite of seventeen diverse English probing tasks and use it to better understand the linguistic knowledge contained within \cwrs.
In contrast to previous studies that analyze the properties and task performance of sentence embeddings \cite[\emph{inter alia}]{adi2016fine,Conneau2018WhatC}, we specifically focus on understanding the \cwrs of individual or pairs of words. 
We release this analysis toolkit to support future work in probing the contents of representations.\footnote{\url{http://nelsonliu.me/papers/contextual-repr-analysis}}
See \appref{probing_task_details} for details about task setup.

\subsection{Token Labeling}
The majority of past work in probing the internal representations of neural models has examined various token labeling tasks, where a decision is made independently for each token \citep[\emph{inter alia}]{Belinkov2017WhatDN,Belinkov2017EvaluatingLO,blevins2018deep}. We synthesize these disparate studies and build upon them by proposing additional probing tasks.

The \textbf{part-of-speech tagging (POS)} task assesses whether \cwrs capture basic syntax.
We experiment with two standard datasets: the Penn Treebank (PTB; \citealp{marcus1993building}) and the Universal Dependencies English Web Treebank (UD-EWT; \citealp{silveira14gold}).

The \textbf{CCG supertagging (CCG)} task assesses the vectors' fine-grained information about the syntactic roles of words in context. It is considered ``{almost parsing}'' \citep{bangalore1999supertagging}, since a sequence of supertags maps a sentence to a small set of possible parses. We use CCGbank \citep{hockenmaier2007ccgbank}, a conversion of the PTB into CCG derivations.

The \textbf{syntactic constituency ancestor tagging} tasks are designed to probe the vectors' knowledge of hierarchical syntax. For a given word, the probing model is trained to predict the constituent label of its parent \textbf{(Parent)}, grandparent \textbf{(GParent)}, or great-grandparent \textbf{(GGParent)} in the phrase-structure tree (from the PTB).

In the \textbf{semantic tagging} task (\textbf{ST}), tokens are assigned labels that reflect their semantic role in context.
These semantic tags assess lexical semantics, and they abstract over redundant POS distinctions and disambiguate useful cases within POS tags. We use the  dataset of \citet{bjerva2016semantic}; the tagset has since been developed as part of the Parallel Meaning Bank \citep{Abzianidze2017ThePM}.

\begin{figure}[t]
	\includegraphics[width=0.95\columnwidth]{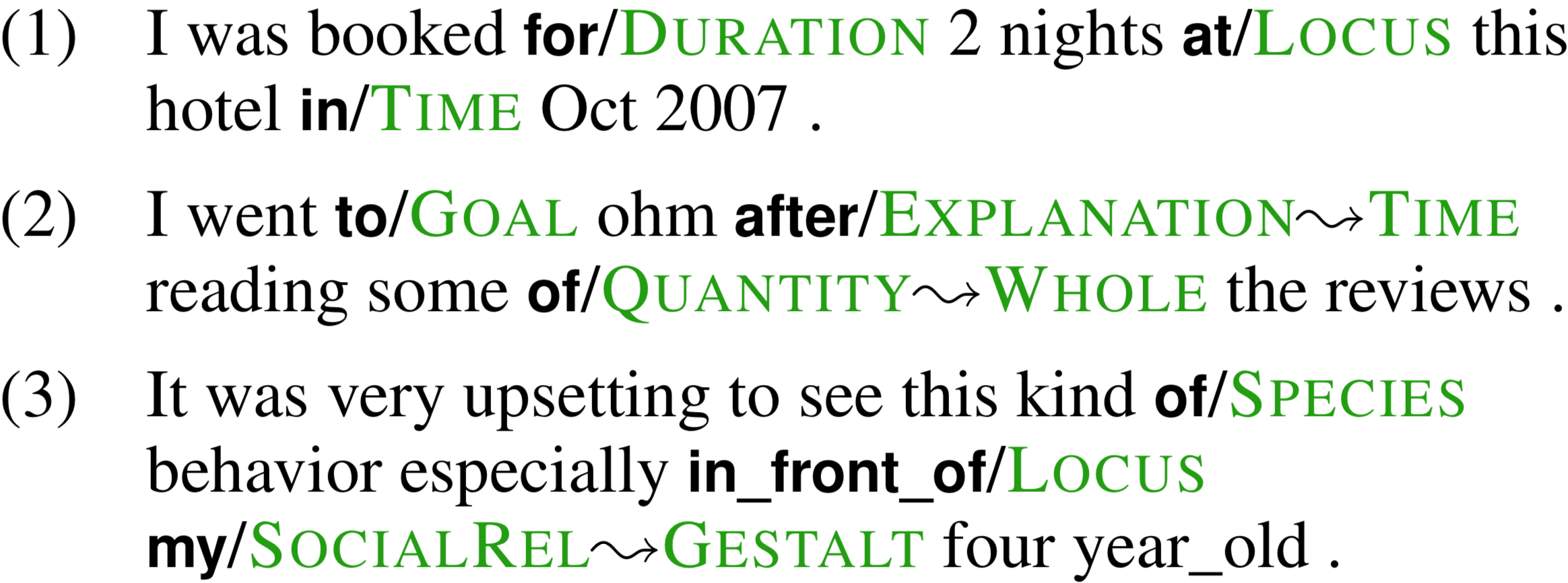}
	\caption{Annotated sentences from the STREUSLE 4.0 corpus, used in the preposition supersense disambiguation task. Prepositions are marked by boldface, immediately followed by their labeled function. If applicable, $\leadsto$ precedes the preposition's labeled role. Figure reproduced from \citet{Schneider2018ComprehensiveSD}.}
\label{fig:preposition_supersense_id_example}
\end{figure}

\textbf{Preposition supersense disambiguation} is the task of classifying a preposition's lexical semantic contribution (the function; \textbf{PS-fxn}) and the semantic role or relation it mediates (the role; \textbf{PS-role}). This task is a specialized kind of word sense disambiguation, and examines one facet of lexical semantic knowledge. In contrast to the tagging tasks above, the model is trained and evaluated on single-token prepositions (rather than making a decision for every token in a sequence). We use the STREUSLE 4.0 corpus  \citep{Schneider2018ComprehensiveSD}; example sentences appear in \Cref{fig:preposition_supersense_id_example}.

The \textbf{event factuality (EF)} task involves labeling phrases with the factuality of the events they describe \citep{Saur2009FactBankAC,Saur2012AreYS,Marneffe2012DidIH}. For instance, in the following example reproduced from \citet{Rudinger2018NeuralMO}, ``(1a) conveys that the leaving didn't happen, while the superficially similar (1b) does not''. 

\ex.
\a. Jo didn't remember to \textbf{leave}.
\b. Jo didn't remember \textbf{leaving}.

We use the Universal Decompositional Semantics It Happened v2 dataset \citep{Rudinger2018NeuralMO}, and the model is trained to predict a (non)factuality value in the range $[-3, 3]$. Unlike the tagging tasks above, this task is treated as a regression problem, where a prediction is made only for tokens corresponding to events (rather than every token in a sequence). Performance is measured using Pearson correlation ($r$); we report ($r \times 100$) so metrics for all tasks fall between 0 and 100.

\subsection{Segmentation}
Several of our probing tasks involve segmentation using BIO or IO tags.  Here the model is trained to predict labels from \emph{only} a single word's \cwr.

\textbf{Syntactic chunking (Chunk)} tests whether \cwrs contain notions of spans and boundaries; the task is to segment text into shallow constituent chunks. We use the CoNLL 2000 shared task dataset \citep{tjong2000introduction}.

\textbf{Named entity recognition (NER)} examines whether \cwrs encode information about entity types. We use the CoNLL 2003 shared task dataset \citep{tjong2003introduction}.

\textbf{Grammatical error detection} \textbf{(GED)} is the task of identifying tokens which need to be edited in order to produce a grammatically correct sentence. Given that \cwrs are extracted from models trained on large amounts of grammatical text, this task assesses whether embeddings encode features that indicate anomalies in their input (in this case, ungrammaticality). We use the First Certificate in English  \citep{yannakoudakis2011new} dataset, converted into sequence-labeling format by \citet{rei2016compositional}.

The \textbf{conjunct identification (Conj)} task challenges the model to identify the tokens that comprise the conjuncts in a coordination construction. Doing so requires highly specific syntactic knowledge. The data comes from the coordination-annotated PTB of \citet{ficler2016coordination}.

\subsection{Pairwise Relations}
We also design probing tasks that examine whether relationships \emph{between} words are encoded in \cwrs. In these tasks, given a word pair $w_1$, $w_2$, we input $[w_1, w_2, w_1 \odot w_2]$ into the probing model; it is trained to predict information about the relation between the tokens \citep{belinkovthesis}.

We distinguish between \textbf{arc prediction} and \textbf{arc classification} tasks. Arc prediction is a binary classification task, where the model is trained to identify \emph{whether} a relation exists between two tokens. %
Arc classification is a multiclass classification task, where the model is provided with two tokens that are linked via some relationship and trained to identify \emph{how} they are related.

For example, in the \textbf{syntactic dependency arc prediction} task, the model is given the representations of two tokens $(w_{a}, w_{b})$ and trained to predict whether the sentence's syntactic dependency parse contains a dependency arc with $w_{a}$ as the head and ${w_{b}}$ as the modifier. The \textbf{syntactic dependency arc classification} task presents the model with the representations of two tokens $(w_{\mathit{head}}, w_{\mathit{mod}})$, where $w_{\mathit{mod}}$ is the modifier of ${w_{\mathit{head}}}$, and the model is trained to predict the type of syntactic relation that link them (the label on that dependency arc). We use the PTB (converted to UD) and the UD-EWT.

Similarly, \textbf{semantic dependency arc prediction} trains the model to predict whether two tokens are connected by a semantic dependency arc, while the \textbf{semantic dependency arc classification} task trains models to classify the semantic relations between tokens. We use the dataset from the SemEval 2015 shared task \citep{oepen2015semeval} with the DELPH-IN MRS-Derived Semantic Dependencies (DM) target representation.

The syntactic and semantic dependency arc prediction and classification tasks are closely related to state-of-the-art models for semantic and syntactic dependency parsing, which score pairs of \cwrs to make head attachment and arc labeling decisions \citep{Dozat2016DeepBA, Dozat2018SimplerBM}.

To generate negative examples for the dependency arc prediction tasks, we take each positive example $(w_{\mathit{head}}, w_{\mathit{mod}})$ and generate a new negative example $(w_{\mathit{rand}}, w_{\mathit{mod}})$. $w_{\mathit{rand}}$ is a random token in the sentence that is not the head of $w_{\mathit{mod}}$. Thus, the datasets used in these tasks are balanced.

We also consider a \textbf{coreference arc prediction} task, where the model is trained to predict whether two entities corefer from their \cwrs. We use the dataset from the CoNLL 2012 shared task \citep{Pradhan2012CoNLL2012ST}. To generate negative examples, we follow a similar procedure as the dependency arc prediction tasks: given a positive example $(w_{a}, w_{b})$, where $w_{b}$ occurs after $w_{a}$ and the two tokens share a coreference cluster, we create a negative example $(w_{\mathit{random\_entity}}, w_{b})$, where $w_{\mathit{random\_entity}}$ is a token that occurs before $w_{b}$ and belongs to a different coreference cluster.

\section{Models}

\paragraph{Probing Model}  We use a linear model as our probing model; limiting its capacity enables us to focus on what information can be \emph{easily} extracted from \cwrs.
 See \appref{probing_model_details} for probing model training hyperparameters and other details.

\paragraph{Contextualizers} We study six publicly-available models for contextualized word representation in English.

\textbf{ELMo} \citep{Peters2018DeepCW} concatenates the output of two contextualizers independently trained on the bidirectional language modeling (biLM) task. \textbf{ELMo (original)} uses a 2-layer LSTM for contextualization. We also study two variations from \citet{Peters2018DissectingCW}: \textbf{ELMo (4-layer)} uses a 4-layer LSTM, and \textbf{ELMo (transformer)} uses a 6-layer transformer \citep{Vaswani2017AttentionIA}. Each of these models is trained on 800M tokens of sentence-shuffled newswire text (the 1 Billion Word Benchmark; \citealp{Chelba2014OneBW}).

The \textbf{OpenAI transformer} \citep{Radford2018IL} is a left-to-right 12-layer transformer language model trained on 800M tokens of contiguous text from over 7,000 unique unpublished books (BookCorpus; \citealp{Zhu2015AligningBA}).

\textbf{BERT} \citep{devlin2018bert} uses a bidirectional transformer jointly trained on a masked language modeling task and a next sentence prediction task. The model is trained on BookCorpus and the English Wikipedia, a total of approximately 3300M tokens. We study \textbf{BERT (base, cased)}, which uses a 12-layer transformer, and \textbf{BERT (large, cased)}, which uses a 24-layer transformer.

\section{Pretrained Contextualizer Comparison}
\label{sec:linguistic_knowledge}

To better understand the linguistic knowledge captured by pretrained contextualizers, we analyze each of their layers with our set of probing tasks.
These contextualizers differ in many respects, and it is outside the scope of this work to control for all differences between them.  We focus on probing the models that are available to us, leaving a more systematic comparison of training regimes and model architectures to future work.

\subsection{Experimental Setup}
Our probing models are trained on the representations produced by the individual layers of each contextualizer.
We also compare to a linear probing model trained on noncontextual vectors (300-dimensional GloVe trained on the cased Common Crawl; \citealp{pennington2014glove}) to assess the gains from contextualization.

\begin{table*}[]
\centering
\setlength\tabcolsep{4pt}
\footnotesize
\begin{tabular}{lccccccccccc}
\toprule
\multirow{2}{*}{Pretrained Representation} & & & \multicolumn{2}{c}{POS} & & & & & \multicolumn{2}{c}{Supersense ID} & \\
\cmidrule{4-5}\cmidrule{10-11}
& {Avg.} & {CCG} & {PTB} & {EWT} & {Chunk} & {NER} & {ST} & {GED} & {PS-Role} & {PS-Fxn} & {EF} \\
\midrule
ELMo (original) best layer & 81.58 & 93.31 & 97.26 & 95.61 & 90.04 & 82.85 & 93.82 & 29.37 & 75.44 & 84.87 & 73.20 \\
ELMo (4-layer) best layer & 81.58 & 93.81 & \textbf{97.31} & 95.60 & 89.78 & 82.06 & \textbf{94.18} & 29.24 & 74.78 & 85.96 & 73.03 \\
ELMo (transformer) best layer & 80.97 & 92.68 & 97.09 & 95.13 & 93.06 & 81.21 & 93.78 & 30.80 & 72.81 & 82.24 & 70.88 \\
OpenAI transformer best layer & 75.01 & 82.69 & 93.82 &	91.28 & 86.06 & 58.14 & 87.81 & 33.10 & 66.23 & 76.97 & 74.03 \\
BERT (base, cased) best layer & 84.09 & 93.67 & 96.95 & 95.21 & 92.64 & 82.71 & 93.72 & 43.30 & \textbf{79.61} & 87.94 & 75.11 \\
BERT (large, cased) best layer & \textbf{85.07} & \textbf{94.28} & 96.73 & \textbf{95.80} & \textbf{93.64} & \textbf{84.44} & 93.83 & \textbf{46.46} & 79.17 & \textbf{90.13} & \textbf{76.25} \\
\midrule
GloVe (840B.300d) & 59.94 & 71.58 & 90.49 &	83.93 & 62.28 & 53.22 & 80.92 & 14.94 & 40.79 & 51.54 & 49.70\\
\midrule
\begin{tabular}{@{}l@{}}Previous state of the art \\ (without pretraining)\end{tabular} & 83.44 & 94.7\enspace & 97.96 & 95.82 & 95.77 & 91.38 & 95.15 & 39.83 & 66.89 & 78.29 & 77.10 \\
\bottomrule
\end{tabular}
\caption{Performance of the best layerwise linear probing model for each contextualizer compared against a GloVe-based linear probing baseline and the previous state of the art. The best contextualizer for each task is bolded. Results for all layers on all tasks, and papers describing the prior state of the art, are given in \appref{all_pretrained_contextualizer_results}.}
\label{tab:glove_vs_contextualizers_vs_sota}
\end{table*}

\subsection{Results and Discussion}
\Cref{tab:glove_vs_contextualizers_vs_sota} compares each contextualizer's best-performing probing model with the GloVe baseline and the previous state of the art for the task (excluding methods that use pretrained \cwrs).\footnote{See \appref{sota_citations} for references to the previous state of the art (without pretraining).}\textsuperscript{,}\footnote{For brevity, in this section we omit probing tasks that cannot be compared to prior work. See \appref{all_pretrained_contextualizer_results} for pretrained contextualizer performance for all layers and all tasks.}

With just a linear model, we can readily extract much of the information needed for high performance on various NLP tasks.
In all cases, \cwrs perform significantly better than the noncontextual baseline.
Indeed, we often see probing models rivaling or exceeding the performance of (often carefully tuned and task-specific) state-of-the-art models. In particular, the linear probing model surpasses the published state of the art for grammatical error detection and preposition supersense identification (both role and function).

Comparing the ELMo-based contextualizers, we see that ELMo (4-layer) and ELMo (original) are essentially even, though both recurrent models outperform ELMo (transformer).
We also see that the OpenAI transformer significantly underperforms the ELMo models and BERT.
Given that it is also the only model trained in a unidirectional (left-to-right) fashion, this reaffirms that bidirectionality is a crucial component for the highest-quality contextualizers \citep{devlin2018bert}.
In addition, the OpenAI transformer is the only model trained on lowercased text, which hinders its performance on tasks like NER.
BERT significantly improves over the ELMo and OpenAI models.

Our probing task results indicate that current methods for \cwr do not capture much transferable information about entities and coreference phenomena in their input (e.g., the NER results in \Cref{tab:glove_vs_contextualizers_vs_sota} and the coreference arc prediction results in \appref{all_pretrained_contextualizer_results}).
To alleviate this weakness, future work could augment pretrained contextualizers with explicit entity representations \citep{Ji2017DynamicER,yang2016reference,Bosselut2017SimulatingAD}.

\paragraph{Probing Failures} While probing models are at or near state-of-the-art performance across a number of tasks, they also do not perform as well on several others, including NER, grammatical error detection, and conjunct identification.
This may occur because (1) the \cwr simply does not encode the pertinent information or any predictive correlates, or (2) the probing model does not have the capacity necessary to extract the information or predictive correlates from the vector. 
In the former case, learning \emph{task-specific contextual features} might be necessary for encoding the requisite task-specific information into the \cwrs. 
Learning task-specific contextual features with a contextual probing model also helps with (2), but we would expect the results to be comparable to increasing the probing model's capacity.

To better understand the failures of our probing model, we experiment with (1) a contextual probing model that uses a task-trained LSTM (unidirectional, 200 hidden units) before the linear output layer (thus adding task-specific contextualization) or (2) replacing the linear probing model with a multilayer perceptron (MLP; adding more parameters to the probing model:  a single 1024d hidden layer activated by ReLU). These alternate probing models have nearly the same number of parameters (LSTM + linear has slightly fewer).

We also compare to a full-featured model to estimate an upper bound on performance for our probing setup.
In this model, the \cwrs are inputs to a 2-layer BiLSTM with 512 hidden units, and the output is fed into a MLP with a single 1024-dimensional hidden layer activated by a ReLU to predict a label. A similar model, augmented with a conditional random field (CRF; \citealp{Lafferty2001ConditionalRF}), achieved state-of-the-art results on the CoNLL 2003 NER dataset \citep{Peters2018DeepCW}. We remove the CRF, since other probing models have no global context.

\begin{table}[]
\centering
\setlength\tabcolsep{4pt}
\footnotesize
\begin{tabular}{lrrrr}
\toprule
Probing Model & NER & GED & Conj & GGParent\\
\midrule
Linear & 82.85 & 29.37 & 38.72 & 67.50 \\
MLP (1024d) & 87.19 & 47.45 & 55.09 & 78.80 \\
LSTM (200d)  + Linear %
& \textbf{88.08} & \textbf{48.90} & \textbf{78.21} & \textbf{84.96} \\
\midrule
\begin{tabular}{@{}l@{}}BiLSTM (512d) \\ + MLP (1024d)\end{tabular} & 90.05 & 48.34 & 87.07 & 90.38 \\
\bottomrule
\end{tabular}
\caption{Comparison of different probing models trained on ELMo (original); best-performing probing model is bolded. Results for each probing model are from the highest-performing contextualizer layer. Enabling probing models to learn task-specific contextual features (with LSTMs) yields outsized benefits in tasks requiring highly specific information.}

\label{tab:adding_parameters_experiment}
\end{table}

For this experiment, we focus on the ELMo (original) pretrained contextualizer.
\Cref{tab:adding_parameters_experiment} presents the performance of the best layer within each alternative probing model on the two tasks with the largest gap between the linear probing model and state-of-the-art methods: NER and grammatical error detection.
We also include great-grandparent prediction and conjunct identification, two tasks that require highly specific syntactic knowledge.
In all cases, we see that adding more parameters (either by replacing the linear model with a MLP, or using a contextual probing model) leads to significant gains over the linear probing model.
On NER and grammatical error detection, we observe very similar performance between the MLP and LSTM + Linear models---this indicates that the probing model simply needed more capacity to extract the necessary information from the \cwrs.
On conjunct identification and great-grandparent prediction, two tasks that probe syntactic knowledge unlikely to be encoded in \cwrs, adding parameters as a task-trained component of our probing model leads to large gains over simply adding parameters to the probing model.
This indicates that the pretrained contextualizers do not capture the information necessary for the task, since such information is learnable by a task-specific contextualizer.

This analysis also reveals insights about contextualizer fine-tuning, which seeks to specialize the \cwrs for an end task \citep{howard2018universal,Radford2018IL,devlin2018bert}.
Our results confirm that task-trained contextualization is important when the end task requires specific information that may not be captured by the pretraining task (\S\ref{sec:linguistic_knowledge}).
However, such end-task--specific contextualization can come from either fine-tuning \cwrs or using fixed output features as inputs to a task-trained contextualizer; \citet{peters2019tune} begins to explore when each approach should be applied.

\section{Analyzing Layerwise Transferability}
\label{sec:layerwise_transferability}

\begin{figure}[t]
\begin{tabular}{c}
(a) ELMo (original) \\
\includegraphics[width=0.9\columnwidth]{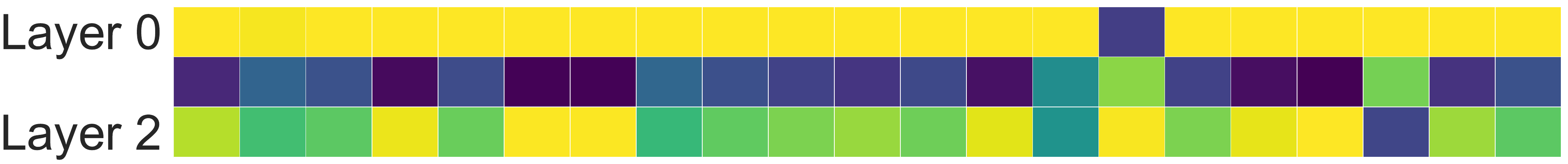} \\
(b) ELMo (4-layer) \\
\includegraphics[width=0.9\columnwidth]{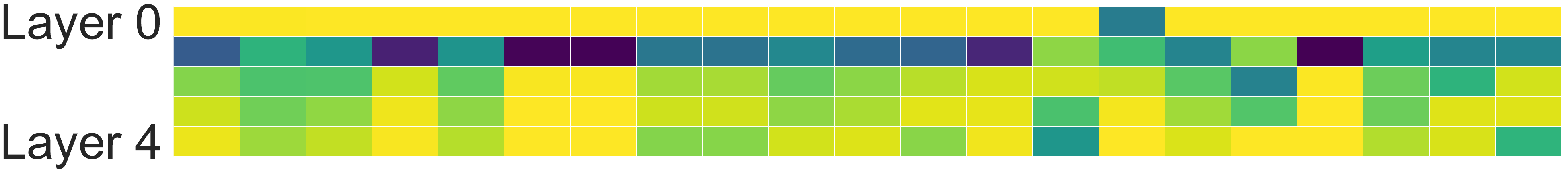} \\
(c) ELMo (transformer) \\
\includegraphics[width=0.9\columnwidth]{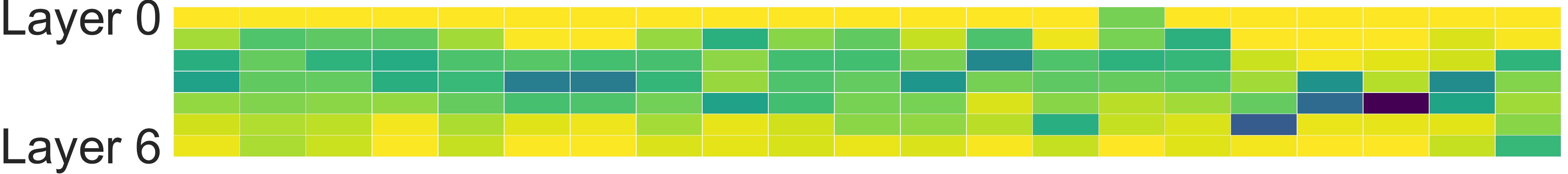} \\
(d) OpenAI transformer \\
\includegraphics[width=0.9\columnwidth]{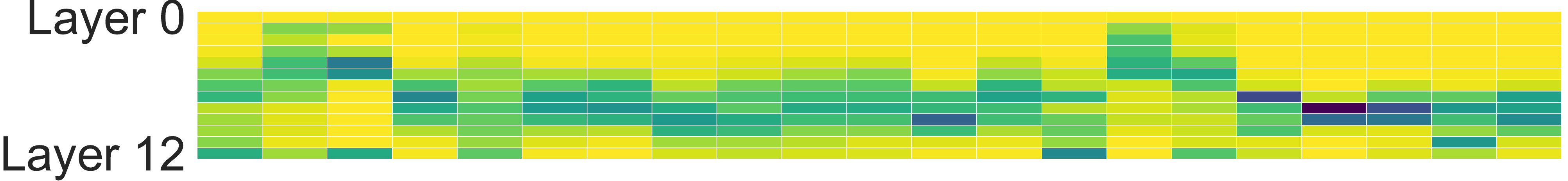} \\
(e) BERT (base, cased) \\
\includegraphics[width=0.9\columnwidth]{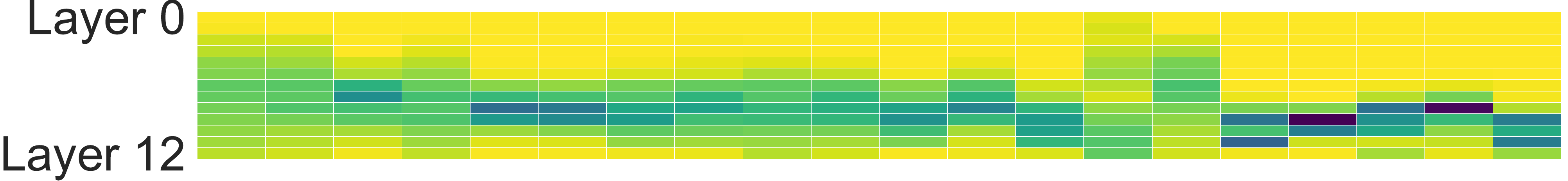} \\
(f) BERT (large, cased) \\
\includegraphics[width=0.9\columnwidth]{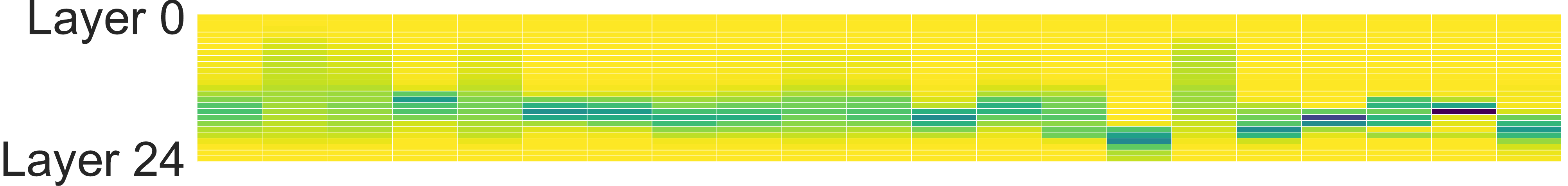} \\
\includegraphics[width=0.9\columnwidth]{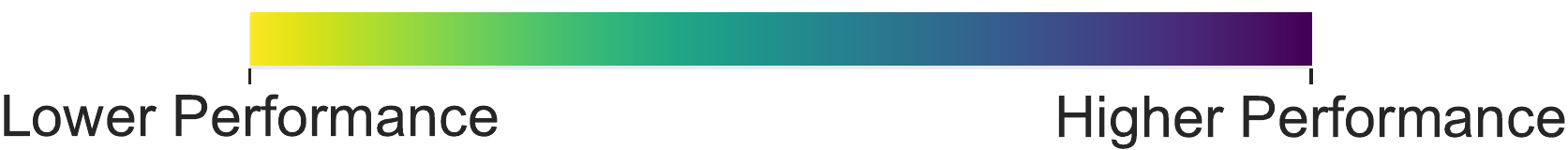} \\
\end{tabular}
  \caption{A visualization of layerwise patterns in task performance. Each column represents a probing task, and each row represents a contextualizer layer.}
  \label{fig:pretrained_layerwise_patterns}
\end{figure}

We quantify the transferability of \cwrs by how well they can do on the range of probing tasks---representations that are more transferable will perform better than alternatives across tasks.
When analyzing the representations produced by each layer of pretrained contextualizers, we observe marked patterns in layerwise transferability (\Cref{fig:pretrained_layerwise_patterns}).
The first layer of contextualization in recurrent models (original and 4-layer ELMo) is consistently the most transferable, even outperforming a scalar mix of layers on most tasks (see \appref{all_pretrained_contextualizer_results} for scalar mix results). \citet{Schuster2019} see the same trend in English dependency parsing.
By contrast, transformer-based contextualizers have no single most-transferable layer; the best performing layer for each task varies, and is usually near the middle. Accordingly, a scalar mix of transformer layers outperforms the best individual layer on most tasks (see \appref{all_pretrained_contextualizer_results}).

Pretraining encourages the model to encode pretraining-task--specific information; they learn transferable features incidentally. We hypothesize that this is an inherent trade-off---since these models used fixed-sized vector representations, task-specificity comes at the cost of generality and transferability. To investigate the task-specificity of the representations generated by each contextualizer layer, we assess how informative each layer of representation is for the pretraining task, essentially treating it as a probe.

\subsection{Experimental Setup}

We focus on the ELMo-based models, since the authors have released code for training their contextualizers. Furthermore, the ELMo-based models facilitate a controlled comparison---they only differ in the contextualizer architecture used.

We evaluate how well \cwr features perform the pretraining task---bidirectional language modeling.
Specifically, we take the pretrained representations for each layer and relearn the language model softmax classifiers used to predict the next and previous token. %
The ELMo models are trained on the Billion Word Benchmark, so we retrain the softmax classifier on similar data to mitigate any possible effects from domain shift. 
We split the held-out portion of the Billion Word Benchmark into  train (80\%, 6.2M tokens) and evaluation (20\%, 1.6M tokens) sets and use this data to retrain and evaluate the softmax classifiers.
We expect that biLM perplexity will be lower when training the softmax classifiers on representations from layers that capture more information about the pretraining task. %

\subsection{Results and Discussion}
\begin{figure}[t]
\begin{tabular}{cc}
(a) ELMo (original) & (b) ELMo (4-layer) \\
\includegraphics[width=0.45\columnwidth]{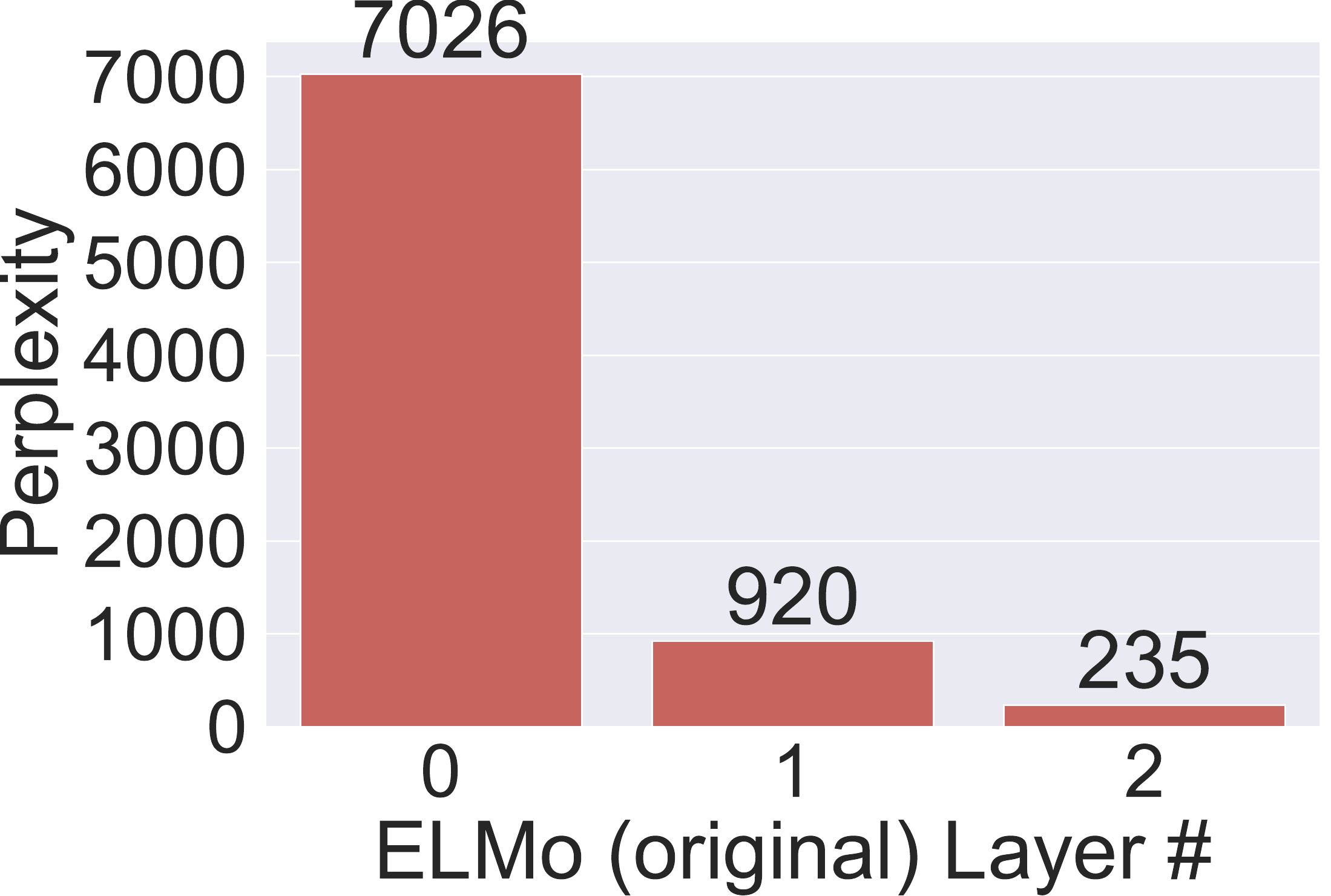} &
\includegraphics[width=0.45\columnwidth]{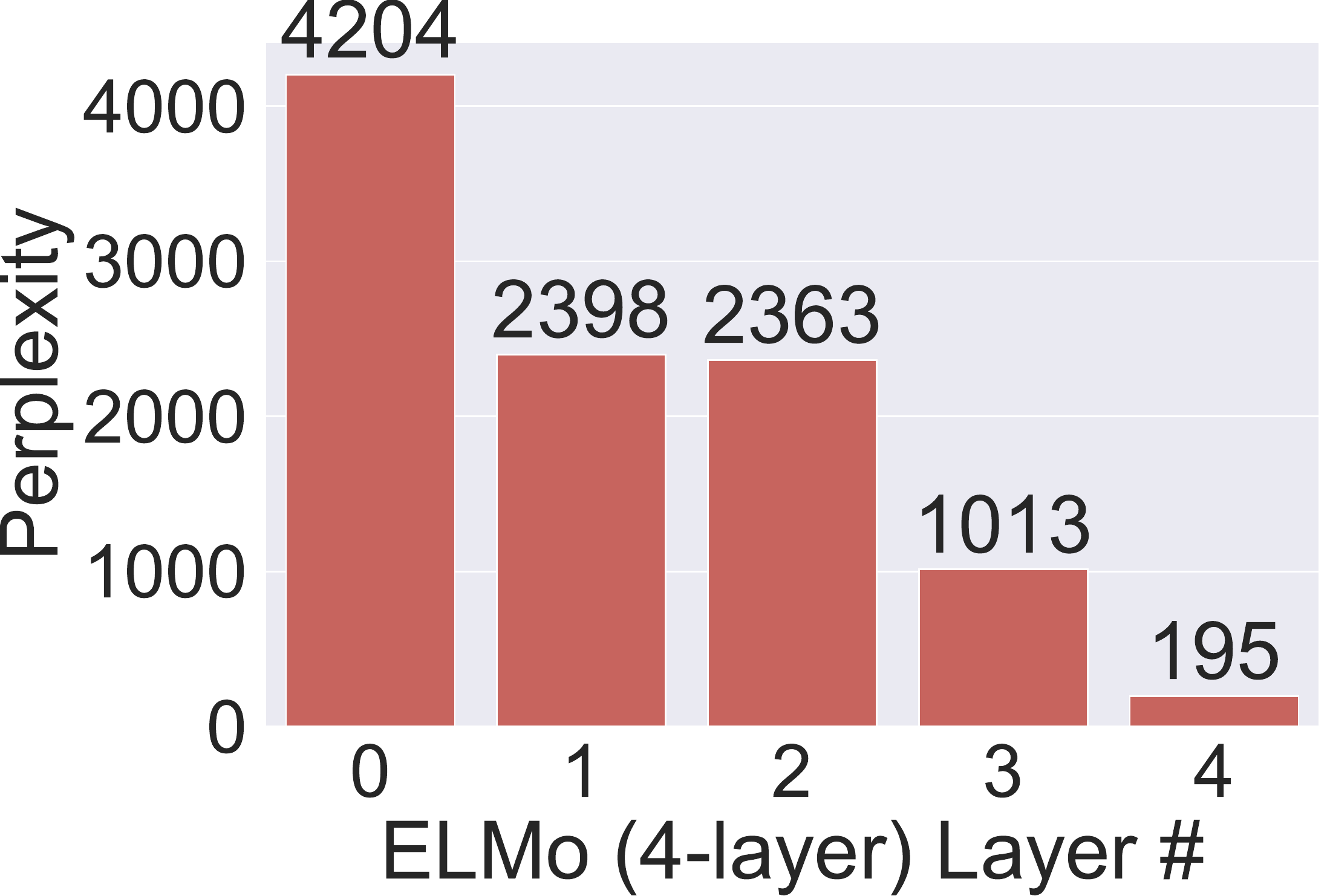} \\
\multicolumn{2}{c}{(c) ELMo (transformer)} \\
\multicolumn{2}{c}{\includegraphics[width=0.9\columnwidth]{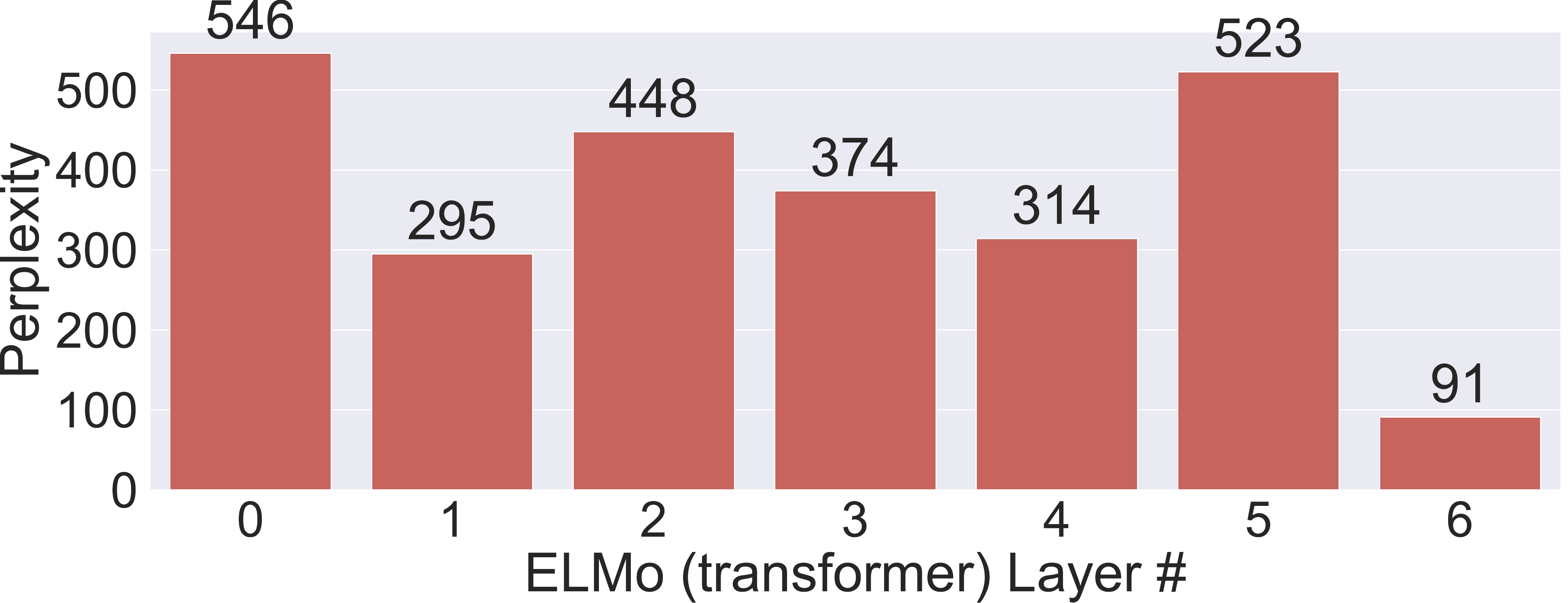}}
\end{tabular}
  \caption{Bidirectional language modeling as a probe: average of forward and backward perplexity (lower is better) of each ELMo contextualizer layer. We see a monotonic decrease in BiLM perplexity when trained on the outputs of higher LSTM layers, but transformer layers do not exhibit the same pattern.}
  \label{fig:layerwise_lm_performance}
\end{figure}

\Cref{fig:layerwise_lm_performance} presents the performance of softmax classifiers trained to perform the bidirectional language modeling task, given just the \cwrs as input. We notice that higher layers in recurrent models consistently achieve lower perplexities. Interestingly, we see that layers 1 and 2 in the 4-layer ELMo model have very similar performance---this warrants further exploration. On the other hand, the layers of the ELMo (transformer) model do not exhibit such a monotonic increase. While the topmost layer is best (which we expected,  since this is the vector originally fed into a softmax classifier during pretraining), the middle layers show varying performance. 
Across all models, the representations that are better-suited for language modeling are also those that exhibit worse probing task performance (\Cref{fig:pretrained_layerwise_patterns}), indicating that contextualizer layers trade off between encoding general and task-specific features.

These results also reveal a difference in the layerwise behavior of LSTMs and transformers; moving up the LSTM layers yields more task-specific representations, but the same does not hold for transformers.
Better understanding the differences between transformers and LSTMs is an active area of research \citep{Chen2018TheBO,Tang2018WhySA}, and we leave further exploration of these observations to future work.

These observations motivate the gradual unfreezing method of \citet{howard2018universal}, where the model layers are progressively unfrozen (starting from the final layer) during the fine-tuning process. Given our observation that higher-level LSTM layers are less general (and more pretraining task-specific), they likely have to be fine-tuned a bit more in order to make them appropriately task specific. Meanwhile, the base layer of the LSTM already learns highly transferable features, and may not benefit from fine-tuning.

\section{Transferring Between  Tasks}
\label{sec:pretraining_task_transferability}

Successful pretrained contextualizers have used self-supervised tasks such as bidirectional language modeling \citep{Peters2018DeepCW} and next sentence prediction \citep{devlin2018bert}, which enable the use of large, unannotated text corpora. However, contextualizers can also be pretrained on explicitly supervised objectives, as done in  pretrained \emph{sentence} embedding methods \citep{Conneau2017SupervisedLO}. To better understand how the choice of pretraining task affects the linguistic knowledge within and transferability of \cwrs, we compare pretraining on a range of different explicitly-supervised tasks with bidirectional language model pretraining.

\subsection{Experimental Setup}

To ensure a controlled comparison of different pretraining tasks, we fix the contextualizer's architecture and pretraining dataset. All of our contextualizers use the ELMo (original) architecture, and the training data from each of the pretraining tasks is taken from the PTB. Each of the (identical) models thus see the same tokens, but the supervision signal differs.\footnote{We omit the OpenAI transformer and BERT from this comparison, since  code for pretraining these contextualizers is not publicly available.}
We compare to (1) a noncontextual baseline (GloVe) to assess the effect of contextualization, (2) a randomly-initialized, untrained ELMo (original) baseline to measure the effect of pretraining, and (3) the ELMo (original) model pretrained on the Billion Word Benchmark to examine the effect of training the bidirectional language model on more data.

\subsection{Results and Discussion}

\Cref{tab:different_pretrain_tasks} presents the average target task performance of each layer in contextualizers pretrained on twelve different tasks (biLM and the eleven tasks from \S\ref{sec:probing_tasks} with PTB annotations).
Bidirectional language modeling pretraining is the most effective on average. However, the settings that achieve the highest performance for individual target tasks often involve transferring between related tasks (not shown in \Cref{tab:different_pretrain_tasks}; see \appref{full_results_transferring_pretraining_tasks}).
For example, when probing \cwrs on the syntactic dependency arc classification (EWT) task, we see the largest gains from pretraining on the task itself, but with a different dataset (PTB). However, pretraining on syntactic dependency arc prediction (PTB), CCG supertagging, chunking, the ancestor prediction tasks, and semantic dependency arc classification all give better performance than bidirectional language model pretraining. 

Although related task transfer is beneficial, we naturally see stronger results from training on more data (the ELMo original BiLM trained on the Billion Word Benchmark). This indicates that the transferability of pretrained \cwrs relies on pretraining on large corpora, emphasizing the utility and importance of self-supervised pretraining.

Furthermore, layer 0 of the BiLM is the highest-performing single layer among PTB-pretrained contextualizers. This observation suggests that lexical information is the source of the language model's initial generalizability, since layer 0 is the output of a character-level convolutional neural network with no token-level contextual information.

\begin{table}[t]
\centering
\setlength\tabcolsep{4pt}
\footnotesize
\begin{tabular}{lcccc}
\toprule
Pretraining Task & \multicolumn{4}{c}{\begin{tabular}{@{}c@{}} Layer Average \\ Target Task Performance \end{tabular}} \\
\cmidrule{2-5}
 & 0 & 1 & 2 & Mix \\
\midrule
CCG & 56.70	& 64.45 & 63.71 & 66.06 \\
Chunk & 54.27 & 62.69 & 63.25 & 63.96 \\
POS & 56.21 & 63.86 & 64.15 & 65.13 \\
Parent & 54.57 & 62.46 & 61.67 & 64.31 \\
GParent & 55.50 & 62.94 & 62.91 & 64.96 \\
GGParent & 54.83 & 61.10 & 59.84 & 63.81 \\
Syn.~Arc Prediction & 53.63 & 59.94 & 58.62 & 62.43 \\
Syn.~Arc Classification & 56.15 & 64.41 & 63.60 & 66.07 \\
Sem.~Arc Prediction & 53.19 & 54.69 & 53.04 & 59.84 \\
Sem.~Arc Classification & 56.28 & 62.41 & 61.47 & 64.67 \\
Conj & 50.24 & 49.93 & 48.42 & 56.92 \\
BiLM & 66.53 & 65.91 & 65.82 & 66.49 \\
\midrule
GloVe (840B.300d) & \multicolumn{4}{c}{60.55}\\
Untrained ELMo (original) & 52.14 & 39.26 & 39.39 & 54.42 \\
\midrule
\begin{tabular}{@{}l@{}}ELMo (original) \\ (BiLM on 1B Benchmark) \end{tabular} & 64.40 & 79.05 & 77.72 & 78.90 \\
\bottomrule
\end{tabular}
\caption{Performance (averaged across target tasks) of contextualizers pretrained on a variety of tasks.}
\label{tab:different_pretrain_tasks}
\end{table}

\section{Related Work}

Methodologically, our work is most similar to \citet{shi2016does}, \citet{adi2016fine}, and \citet{Hupkes2018VisualisationA}, who use the internal representations of neural models to predict properties of interest. \citet{Conneau2018WhatC} construct probing tasks to study the linguistic properties of sentence embedding methods. We focus on contextual word representations, which have achieved state-of-the-art results on a variety of tasks, and examine a broader range of linguistic knowledge.

In contemporaneous work, \citet{Tenney2019What} evaluate CoVe \citep{mccann2017learned},  ELMo \citep{Peters2018DeepCW}, the OpenAI Transformer \citep{Radford2018IL}, and BERT \citep{devlin2018bert} on a variety of sub-sentence linguistic analysis tasks. Their results also suggest that the aforementioned pretrained models for contextualized word representation encode stronger notions of syntax than higher-level semantics. They also find that using a scalar mix of output layers is particularly effective in deep transformer-based models, aligned with our own probing results and our observation that transformers tend to encode transferable features in their intermediate layers. Furthermore, they find that ELMo's performance cannot be explained by a model with access to only local context, indicating that ELMo encodes linguistic features from distant tokens.

Several other papers have examined how architecture design and choice of pretraining task affect the quality of learned \cwrs.
\citet{Peters2018DissectingCW} study how the choice of neural architecture influences the end-task performance and qualitative properties of \cwrs derived from bidirectional language models (ELMo).
\citet{bowman2018looking} compare a variety of pretraining tasks and explore the the impact of multitask learning.

Prior work has employed a variety of other methods to study the learned representations in neural models, such as directly examining the activations of individual neurons \citep[\emph{inter alia}]{karpathy2015visualizing,li2015visualizing,shi2016neural}, ablating components of the model and dataset \citep{kuncoro2016recurrent,gaddy2018s,khandelwal2018lm}, or interpreting attention mechanisms \citep{bahdanau2014neural}; see \citet{belinkov2019analysis} for a recent survey. One particularly relevant line of work involves the construction of synthetic tasks that a model can only solve if it captures a particular phenomenon \citep[\emph{inter alia}]{Linzen2016AssessingTA,jumelet2018language,wilcox2018rnn,futrell2019rnns}.
\citet{Zhang2018LanguageMT} compare the syntactic knowledge of language models and neural machine translation systems. We widen the range of pretraining tasks and target probing model tasks to gain a more complete picture. We also focus on a stronger contextualizer architecture, ELMo (original), that has produced state-of-the-art results.

Several studies have sought to intrinsically evaluate noncontextual word representations with word similarity tasks, such as analogies \citep{Mikolov2013DistributedRO}. These methods differ from our approach in that they require no extra parameters and directly assess the vectors, while our probing models must be trained. In this regard, our method is similar to \textsc{qvec} \citep{Tsvetkov2015EvaluationOW}.

\section{Conclusion}

We study the linguistic knowledge and transferability of contextualized word representations with a suite of seventeen diverse probing tasks.
The features generated by pretrained contextualizers are sufficient for high performance on a broad set of tasks.
For tasks that require specific information not captured by the contextual word representation, we show that learning task-specific contextual features helps encode the requisite knowledge.
In addition, our analysis of patterns in the transferability of contextualizer layers shows that the lowest layer of LSTMs encodes the most transferable features, while transformers' middle layers are most transferable.
We find that higher layers in LSTMs are more task-specific (and thus less general), while transformer layers do not exhibit this same monotonic increase in task-specificity.
Prior work has suggested that higher-level contextualizer layers may be expressly encoding higher-level semantic information. Instead, it seems likely that certain high-level semantic phenomena are incidentally useful for the contextualizer's pretraining task, leading to their presence in higher layers.
Lastly, we find that bidirectional language model pretraining yields representations that are more transferable \emph{in general} than eleven other candidate pretraining tasks.

\section*{Acknowledgments}

We thank Johannes Bjerva for sharing the semantic tagging dataset used in \citet{bjerva2016semantic}. We also thank the members of the Noah's ARK group at the University of Washington, the researchers at the Allen Institute for Artificial Intelligence, and the anonymous reviewers for their valuable feedback. NL is supported by a Washington Research Foundation Fellowship and a Barry M. Goldwater Scholarship. YB is supported by the Harvard Mind, Brain, and Behavior Initiative. 

\bibliography{naaclhlt2019}
\bibliographystyle{acl_natbib}

\newpage
\clearpage

\begin{appendices}

\labeledsection{Probing Task Setup Details}{probing_task_details}

\paragraph{Syntactic Constituency Ancestor Tagging} We remove the top-level ROOT node in each sentence. For words that do not have a parent, grandparent, or great-grandparent, we set the label to "None". The example is then treated as any other, and the probing model is required to predict this "None" label during training and evaluation.

\paragraph{Preposition Supersense Disambiguation} Since we focus on the linguistic knowledge within individual or pairs of \cwrs, we train and evaluate our probing models on only single-word adpositions.

\paragraph{Conjunct Identification} Our probing models are only trained and evaluated on sentences with a coordination construction in them.

\labeledsection{Probing Model Training Details}{probing_model_details}

Our probing models are trained with Adam \citep{Kingma2014AdamAM}, using a learning rate of 0.001. 
We train for 50 epochs, using early stopping with a patience of 3.
Our models are implemented in the AllenNLP framework \citep{Gardner2018AllenNLPAD}.

For contextualizers that use subword representations (e.g., the OpenAI transformer and BERT), we aggregate subword representations into token representations by taking a token's representation to be the representation of its final subword.

\labeledsection{References to State-of-the-Art Task-Specific Models (Without Pretraining)}{sota_citations}

\begin{table}[h]
\centering
\footnotesize
\begin{tabular}{ll}
\toprule
Task & \begin{tabular}{@{}l@{}}Previous state of the art \\ (without pretraining)\end{tabular} \\
\midrule
CCG & 94.7 \enspace\citep{lewis2016lstm} \\
POS (PTB) & 97.96 \citep{Bohnet2018MorphosyntacticTW} \\
POS (EWT) & 95.82 \citep{Yasunaga2018RobustMP} \\
Chunk & 95.77 \citep{Hashimoto2017AJM} \\
NER & 91.38 \citep{Hashimoto2017AJM} \\
ST & 95.15 \citep{bjerva2016semantic} \\
GED & 39.83 \citep{Rei2019JointlyLT} \\
PS-Role & 66.89 \citep{Schneider2018ComprehensiveSD} \\
PS-Fxn & 78.29 \citep{Schneider2018ComprehensiveSD} \\
EG & 77.10 \citep{Rudinger2018NeuralMO} \\
\bottomrule
\end{tabular}
\caption{Performance of prior state of the art models (without pretraining) for each task.}
\label{tab:sota_papers}
\end{table}

Note that the performance reported in this paper for the preposition supersense identification models of \citet{Schneider2018ComprehensiveSD} differs from their published result. Their published result is the accuracy on all adpositions; since we only train and evaluate our model on single-word adpositions, the number we report in this paper is the performance of the \citet{Schneider2018ComprehensiveSD} model on only single-word adpositions.

\clearpage
\newpage
\onecolumn
\labeledsection{Performance of Pretrained Contextualizers on All Tasks}{all_pretrained_contextualizer_results}

\subsection{Token Labeling (ELMo and OpenAI Transformer)}

\begin{table*}[h]
\centering
\setlength\tabcolsep{4pt}
\footnotesize
\begin{tabular}{lcccccccccc}
\toprule
\multirow{2}{*}{Pretrained Representation} & & \multicolumn{2}{c}{POS} & & & & & \multicolumn{2}{c}{Supersense ID} & \\
\cmidrule{3-4}\cmidrule{9-10}
& {CCG} & {PTB} & {EWT} & {Parent} & {GParent} & {GGParent} & {ST} & {PS-Role} & {PS-Fxn} & {EF} \\
\midrule
ELMo Original, Layer 0 & 73.43 & 93.31 & 89.71 & 85.23 & 54.58 & 41.57 & 83.99 & 41.45 & 52.41 & 52.49 \\
ELMo Original, Layer 1 & 93.31 & 97.26 & 95.61 & 95.56 & 81.61 & 67.50 & 93.82 & 74.12 & 84.87 & 73.20 \\ 
ELMo Original, Layer 2 & 91.23 & 96.45 & 94.52 & 94.35 & 76.22 & 62.32 & 92.41 & 75.44 & 83.11 & 72.11 \\
ELMo Original, Scalar Mix & 92.96 & 97.19 & 95.09 & 95.56 & 81.56 & 67.42 & 93.86 & 74.56 & 84.65 & 72.96 \\
\midrule
ELMo (4-layer), Layer 0 & 73.41 & 93.42 & 89.30 & 85.45 & 55.40 & 42.22 & 83.95 & 40.13 & 55.26 & 53.58 \\
ELMo (4-layer), Layer 1 & 93.81 & 97.31 & 95.60 & 95.70 & 81.57 & 67.66 & 94.18 & 74.78 & 85.96 & 73.03 \\
ELMo (4-layer), Layer 2 & 92.47 & 97.09 & 95.08 & 95.01 & 77.08 & 63.04 & 93.43 & 74.12 & 85.53 & 70.97 \\
ELMo (4-layer), Layer 3 & 91.56 & 96.82 & 94.56 & 94.65 & 75.58 & 61.04 & 92.82 & 74.12 & 83.55 & 70.66 \\
ELMo (4-layer), Layer 4 & 90.67 & 96.44 & 93.99 & 94.24 & 75.70 & 61.45 & 91.90 & 73.46 & 83.77 & 72.59 \\
ELMo (4-layer), Scalar Mix & 93.23 & 97.34 & 95.14 & 95.55 & 81.36 & 67.47 & 94.05 & 76.10 & 84.65 & 72.70 \\
\midrule
ELMo (transformer), Layer 0 & 73.06 & 93.27 & 89.42 & 85.59 & 55.03 & 41.38 & 83.81 & 41.45 & 54.39 & 53.13 \\
ELMo (transformer), Layer 1 & 91.66 & 97.09 & 94.78 & 94.43 & 77.28 & 62.69 & 93.78 & 65.13 & 80.04 & 67.19 \\
ELMo (transformer), Layer 2 & 92.68 & 96.93 & 95.13 & 95.15 & 81.37 & 67.39 & 93.71 & 69.74 & 80.26 & 70.88 \\
ELMo (transformer), Layer 3 & 92.82 & 96.97 & 94.74 & 95.28 & 82.16 & 68.06 & 93.45 & 70.61 & 82.24 & 70.24 \\
ELMo (transformer), Layer 4 & 91.86 & 96.71 & 94.41 & 94.97 & 81.48 & 67.33 & 92.82 & 72.81 & 82.02 & 69.97 \\
ELMo (transformer), Layer 5 & 91.06 & 96.24 & 93.85 & 94.30 & 79.65 & 64.92 & 91.92 & 69.52 & 79.82 & 70.21 \\
ELMo (transformer), Layer 6 & 90.19 & 96.33 & 93.62 & 93.98 & 77.40 & 63.49 & 91.78 & 65.57 & 80.48 & 70.82 \\
ELMo (transformer), Scalar Mix & 93.66 & 97.35 & 94.59 & 95.16 & 83.38 & 69.29 & 94.26 & 72.59 & 82.46 & 71.94 \\
\midrule
OpenAI transformer, Layer 0 & 71.58 & 89.54 & 87.44 & 84.50 & 56.24 & 46.31 & 81.18 & 37.72 & 48.90 & 55.03 \\
OpenAI transformer, Layer 1 & 78.08 & 93.32 & 89.93 & 88.75 & 63.59 & 53.28 & 85.73 & 43.64 & 61.40 & 63.13 \\ 
OpenAI transformer, Layer 2 & 78.19 & 92.71 & 85.27 & 88.22 & 65.85 & 56.34 & 85.54 & 52.41 & 66.45 & 65.69 \\
OpenAI transformer, Layer 3 & 79.53 & 93.43 & 89.67 & 88.73 & 67.34 & 58.10 & 86.17 & 53.51 & 70.18 & 68.39 \\
OpenAI transformer, Layer 4 & 80.95 & 93.82 & 91.28 & 90.07 & 69.34 & 60.74 & 87.34 & 58.55 & 71.27 & 69.82 \\
OpenAI transformer, Layer 5 & 82.03 & 93.82 & 91.11 & 90.51 & 71.41 & 62.69 & 87.81 & 60.75 & 73.46 & 70.92 \\
OpenAI transformer, Layer 6 & 82.38 & 93.45 & 88.09 & 90.32 & 72.10 & 63.68 & 87.46 & 64.04 & 74.12 & 72.08 \\
OpenAI transformer, Layer 7 & 82.61 & 93.25 & 86.50 & 90.71 & 72.60 & 63.69 & 86.49 & 65.13 & 76.32 & 73.87 \\
OpenAI transformer, Layer 8 & 81.43 & 92.10 & 86.66 & 91.00 & 72.66 & 64.01 & 86.65 & 66.23 & 76.97 & 73.86 \\
OpenAI transformer, Layer 9 & 81.73 & 91.99 & 86.60 & 90.84 & 72.34 & 63.72 & 86.19 & 66.01 & 76.54 & 74.03 \\
OpenAI transformer, Layer 10 & 81.73 & 92.05 & 86.37 & 90.74 & 71.41 & 62.45 & 86.22 & 63.38 & 75.88 & 73.30 \\
OpenAI transformer, Layer 11 & 81.97 & 91.64 & 86.62 & 90.43 & 70.48 & 60.84 & 85.91 & 63.16 & 76.97 & 71.99 \\
OpenAI transformer, Layer 12 & 82.69 & 92.18 & 90.87 & 90.89 & 69.14 & 58.74 & 87.43 & 63.60 & 75.66 & 71.34 \\
OpenAI transformer, Scalar Mix & 83.94 & 94.63 & 92.60 & 92.08 & 73.11 & 64.64 & 88.73 & 64.69 & 79.17 & 74.25 \\
\midrule
GloVe (840B.300d) & 71.58 & 90.49 & 83.93 & 81.77 & 54.01 & 41.21 & 80.92 & 40.79 & 51.54 & 49.70 \\
\midrule
Previous state of the art & 94.7\enspace & 97.96 & 96.73 & - & - & - & 95.15 & 66.89 & 78.29 & 77.10 \\
\bottomrule
\end{tabular}
\caption{Token labeling task performance of a linear probing model trained on top of the ELMo and OpenAI contextualizers, compared against a GloVe-based probing baseline and the previous state of the art.}
\end{table*}

\newpage 

\subsection{Token Labeling (BERT)}

\begin{table*}[h]
\centering
\setlength\tabcolsep{4pt}
\footnotesize
\begin{tabular}{lcccccccccc}
\toprule
\multirow{2}{*}{Pretrained Representation} & & \multicolumn{2}{c}{POS} & & & & & \multicolumn{2}{c}{Supersense ID} & \\
\cmidrule{3-4}\cmidrule{9-10}
& {CCG} & {PTB} & {EWT} & {Parent} & {GParent} & {GGParent} & {ST} & {PS-Role} & {PS-Fxn} & {EF} \\
\midrule
BERT (base, cased), Layer 0 & 71.45 & 89.99 & 86.77 & 84.41 & 55.92 & 46.07 & 82.25 & 42.11 & 54.82 & 52.70 \\
BERT (base, cased), Layer 1 & 81.67 & 93.80 & 90.58 & 89.47 & 62.92 & 50.93 & 88.89 & 50.88 & 67.76 & 59.83 \\
BERT (base, cased), Layer 2 & 88.43 & 95.76 & 93.72 & 92.98 & 71.73 & 57.84 & 92.23 & 63.60 & 75.00 & 64.91 \\
BERT (base, cased), Layer 3 & 89.77 & 96.08 & 94.30 & 93.92 & 73.24 & 58.57 & 92.85 & 64.69 & 78.95 & 65.58 \\
BERT (base, cased), Layer 4 & 91.41 & 96.57 & 94.58 & 94.67 & 76.09 & 61.17 & 93.38 & 66.23 & 79.17 & 67.55 \\
BERT (base, cased), Layer 5 & 92.22 & 96.68 & 94.93 & 95.10 & 77.79 & 63.56 & 93.47 & 68.20 & 82.89 & 69.08 \\
BERT (base, cased), Layer 6 & 93.14 & 96.95 & 95.15 & 95.46 & 79.75 & 65.36 & 93.72 & 76.10 & 84.65 & 71.26 \\
BERT (base, cased), Layer 7 & 93.51 & 96.92 & 95.12 & 95.70 & 80.38 & 65.96 & 93.62 & 77.85 & 86.40 & 71.54 \\
BERT (base, cased), Layer 8 & 93.67 & 96.80 & 95.21 & 95.60 & 81.04 & 66.66 & 93.37 & 79.61 & 87.94 & 73.49 \\
BERT (base, cased), Layer 9 & 93.51 & 96.68 & 94.94 & 95.64 & 80.70 & 66.53 & 93.18 & 79.39 & 86.84 & 75.11 \\
BERT (base, cased), Layer 10 & 93.25 & 96.54 & 94.51 & 95.26 & 79.60 & 65.49 & 92.90 & 79.17 & 86.18 & 74.70 \\
BERT (base, cased), Layer 11 & 92.75 & 96.40 & 94.31 & 95.00 & 78.50 & 64.34 & 92.64 & 77.41 & 85.53 & 75.11 \\
BERT (base, cased), Layer 12 & 92.21 & 96.09 & 93.86 & 94.55 & 76.95 & 62.87 & 92.34 & 78.07 & 84.65 & 73.77 \\
BERT (base, cased), Scalar Mix & 93.78 & 97.02 & 95.63 & 95.83 & 81.67 & 67.48 & 93.85 & 78.51 & 85.96 & 74.88 \\
\midrule
BERT (large, cased), Layer 0 & 71.06 & 89.84 & 86.81 & 84.28 & 55.84 & 46.17 & 82.31 & 38.38 & 54.61 & 52.81 \\
BERT (large, cased), Layer 1 & 79.49 & 92.58 & 89.45 & 88.50 & 60.96 & 49.88 & 87.16 & 53.51 & 65.13 & 59.49 \\
BERT (large, cased), Layer 2 & 83.30 & 94.03 & 91.70 & 90.48 & 64.91 & 51.94 & 89.47 & 58.55 & 71.93 & 62.49 \\
BERT (large, cased), Layer 3 & 83.32 & 94.09 & 91.92 & 90.76 & 64.99 & 52.26 & 89.67 & 58.33 & 72.81 & 62.52 \\
BERT (large, cased), Layer 4 & 88.51 & 95.61 & 93.36 & 93.26 & 70.99 & 56.22 & 92.58 & 65.35 & 78.29 & 65.06 \\
BERT (large, cased), Layer 5 & 89.69 & 95.95 & 94.15 & 93.94 & 72.62 & 57.58 & 93.05 & 62.06 & 76.97 & 65.79 \\
BERT (large, cased), Layer 6 & 90.91 & 96.14 & 94.35 & 94.47 & 75.59 & 60.80 & 93.35 & 62.72 & 78.51 & 67.00 \\
BERT (large, cased), Layer 7 & 91.72 & 96.30 & 94.64 & 94.55 & 76.35 & 60.98 & 93.55 & 67.98 & 81.36 & 66.42 \\
BERT (large, cased), Layer 8 & 91.56 & 96.36 & 94.80 & 94.61 & 76.40 & 61.93 & 93.50 & 66.89 & 80.26 & 68.56 \\
BERT (large, cased), Layer 9 & 91.76 & 96.31 & 94.86 & 94.70 & 75.95 & 61.60 & 93.44 & 66.89 & 82.02 & 69.12 \\
BERT (large, cased), Layer 10 & 91.71 & 96.27 & 94.89 & 94.88 & 75.84 & 61.44 & 93.42 & 68.64 & 79.39 & 69.37 \\
BERT (large, cased), Layer 11 & 92.01 & 96.26 & 94.96 & 95.10 & 77.01 & 62.79 & 93.39 & 70.83 & 81.80 & 71.12 \\
BERT (large, cased), Layer 12 & 92.82 & 96.48 & 95.27 & 95.31 & 78.66 & 64.51 & 93.61 & 74.34 & 84.21 & 72.44 \\
BERT (large, cased), Layer 13 & 93.48 & 96.73 & 95.56 & 95.72 & 80.51 & 65.85 & 93.83 & 76.54 & 85.75 & 72.91 \\
BERT (large, cased), Layer 14 & 93.85 & 96.73 & 95.54 & 95.98 & 81.89 & 67.02 & 93.81 & 78.95 & 87.94 & 72.72 \\
BERT (large, cased), Layer 15 & 94.21 & 96.72 & 95.80 & 96.10 & 82.46 & 67.53 & 93.76 & 79.17 & 89.25 & 72.79 \\
BERT (large, cased), Layer 16 & 94.28 & 96.67 & 95.62 & 96.05 & 82.78 & 67.90 & 93.61 & 78.73 & 90.13 & 74.27 \\
BERT (large, cased), Layer 17 & 94.13 & 96.53 & 95.55 & 95.92 & 82.56 & 67.74 & 93.45 & 79.17 & 87.06 & 75.52 \\
BERT (large, cased), Layer 18 & 93.76 & 96.38 & 95.45 & 95.57 & 81.47 & 67.11 & 93.21 & 79.17 & 87.06 & 75.95 \\
BERT (large, cased), Layer 19 & 93.36 & 96.25 & 95.30 & 95.38 & 80.47 & 66.08 & 93.01 & 76.10 & 85.96 & 76.25 \\
BERT (large, cased), Layer 20 & 93.06 & 96.10 & 94.96 & 95.20 & 79.32 & 64.86 & 92.78 & 78.29 & 87.72 & 75.92 \\
BERT (large, cased), Layer 21 & 91.83 & 95.38 & 94.05 & 94.16 & 76.84 & 62.43 & 91.65 & 74.12 & 82.89 & 75.16 \\
BERT (large, cased), Layer 22 & 89.66 & 93.88 & 92.30 & 92.62 & 74.73 & 60.76 & 89.42 & 73.90 & 82.02 & 74.28 \\
BERT (large, cased), Layer 23 & 88.70 & 93.02 & 91.90 & 92.36 & 73.33 & 59.27 & 88.92 & 69.08 & 80.70 & 73.54 \\
BERT (large, cased), Layer 24 & 87.65 & 92.60 & 90.84 & 91.81 & 71.98 & 57.95 & 88.26 & 69.74 & 78.73 & 72.65 \\
BERT (large, cased), Scalar Mix & 94.48 & 97.17 & 96.05 & 96.27 & 83.51 & 68.90 & 93.96 & 78.95 & 87.06 & 76.13 \\
\bottomrule
\end{tabular}
\caption{Token labeling task performance of a linear probing model trained on top of the BERT contextualizers.}
\end{table*}

\newpage 

\subsection{Segmentation (ELMo and OpenAI Transformer)}

\begin{table*}[h]
\centering
\setlength\tabcolsep{4pt}
\footnotesize
\begin{tabular}{lcccc}
\toprule
Pretrained Representation & Chunk & NER & GED & Conj \\
\midrule
ELMo Original, Layer 0 &  70.68 &  64.39 &  18.49 &  15.59 \\
ELMo Original, Layer 1 &  90.04 &  82.85 &  29.37 &  38.72 \\
ELMo Original, Layer 2 &  86.47 &  82.80 &  26.08 &  29.08 \\
ELMo Original, Scalar Mix &  89.29 &  82.90 &  27.54 &  39.57 \\
\midrule
ELMo (4-layer), Layer 0 &  70.57 &  63.96 &   8.46 &  15.15 \\
ELMo (4-layer), Layer 1 &  89.78 &  81.04 &  28.07 &  36.37 \\
ELMo (4-layer), Layer 2 &  87.18 &  80.19 &  29.24 &  31.44 \\
ELMo (4-layer), Layer 3 &  86.20 &  81.56 &  28.51 &  28.57 \\
ELMo (4-layer), Layer 4 &  85.07 &  82.06 &  23.85 &  26.31 \\
ELMo (4-layer), Scalar Mix &  86.67 &  82.37 &  30.46 &  28.42 \\
\midrule
ELMo (transformer), Layer 0 &  71.01 &  64.23 &  13.25 &  15.69 \\
ELMo (transformer), Layer 1 &  91.75 &  78.51 &  25.29 &  26.56 \\
ELMo (transformer), Layer 2 &  92.18 &  80.92 &  28.63 &  34.99 \\
ELMo (transformer), Layer 3 &  92.14 &  80.80 &  29.16 &  38.23 \\
ELMo (transformer), Layer 4 &  91.32 &  80.47 &  29.71 &  38.52 \\
ELMo (transformer), Layer 5 &  89.18 &  81.21 &  30.80 &  35.49 \\
ELMo (transformer), Layer 6 &  87.96 &  79.77 &  27.20 &  29.17 \\
ELMo (transformer), Scalar Mix &  92.08 &  81.68 &  26.56 &  38.45 \\
\midrule
OpenAI transformer, Layer 0 &  66.59 &  46.29 &  14.78 &  16.84 \\
OpenAI transformer, Layer 1 &  77.87 &  48.88 &  19.72 &  17.59 \\
OpenAI transformer, Layer 2 &  79.67 &  52.13 &  21.59 &  20.72 \\
OpenAI transformer, Layer 3 &  80.78 &  52.40 &  22.58 &  22.36 \\
OpenAI transformer, Layer 4 &  82.95 &  54.62 &  25.61 &  23.04 \\
OpenAI transformer, Layer 5 &  84.67 &  56.25 &  29.69 &  25.53 \\
OpenAI transformer, Layer 6 &  85.46 &  56.46 &  30.69 &  27.25 \\
OpenAI transformer, Layer 7 &  86.06 &  57.73 &  33.10 &  30.68 \\
OpenAI transformer, Layer 8 &  85.75 &  56.50 &  32.17 &  33.06 \\
OpenAI transformer, Layer 9 &  85.40 &  57.31 &  31.90 &  32.65 \\
OpenAI transformer, Layer 10 &  84.52 &  57.32 &  32.08 &  30.27 \\
OpenAI transformer, Layer 11 &  83.00 &  56.94 &  30.22 &  26.60 \\
OpenAI transformer, Layer 12 &  82.44 &  58.14 &  30.81 &  25.19 \\
OpenAI transformer, Scalar Mix &  87.44 &  59.39 &  34.54 &  31.65 \\
\midrule
GloVe (840B.300d) & 62.28 & 53.22 & 14.94 & 10.53  \\
\midrule
Previous state of the art & 95.77 & 91.38 & 34.76 & -  \\
\bottomrule
\end{tabular}
\caption{Segmentation task performance of a linear probing model trained on top of the ELMo and OpenAI contextualizers, compared against a GloVe-based probing baseline and the previous state of the art.}
\end{table*}

\newpage

\subsection{Segmentation (BERT)}

\begin{table*}[h]
\centering
\setlength\tabcolsep{4pt}
\footnotesize
\begin{tabular}{lcccc}
\toprule
Pretrained Representation & Chunk & NER & GED & Conj \\
\midrule
BERT (base, cased), Layer 0 &  69.86 &  53.50 &  12.63 &  16.24 \\
BERT (base, cased), Layer 1 &  75.56 &  66.94 &  16.85 &  21.83 \\
BERT (base, cased), Layer 2 &  86.64 &  71.08 &  22.66 &  22.87 \\
BERT (base, cased), Layer 3 &  87.70 &  73.83 &  25.80 &  25.50 \\
BERT (base, cased), Layer 4 &  90.64 &  77.28 &  31.35 &  29.39 \\
BERT (base, cased), Layer 5 &  91.21 &  78.81 &  32.34 &  30.58 \\
BERT (base, cased), Layer 6 &  92.29 &  80.81 &  37.85 &  35.26 \\
BERT (base, cased), Layer 7 &  92.64 &  81.50 &  40.14 &  35.86 \\
BERT (base, cased), Layer 8 &  92.11 &  82.45 &  42.08 &  42.26 \\
BERT (base, cased), Layer 9 &  91.95 &  82.71 &  43.20 &  43.93 \\
BERT (base, cased), Layer 10 &  91.30 &  82.66 &  42.46 &  43.38 \\
BERT (base, cased), Layer 11 &  90.71 &  82.42 &  43.30 &  41.35 \\
BERT (base, cased), Layer 12 &  89.38 &  80.64 &  39.87 &  39.34 \\
BERT (base, cased), Scalar Mix &  92.96 &  82.43 &  43.22 &  43.15 \\
\midrule
BERT (large, cased), Layer 0 &  70.42 &  53.95 &  13.44 &  16.65 \\
BERT (large, cased), Layer 1 &  73.98 &  65.92 &  16.20 &  19.58 \\
BERT (large, cased), Layer 2 &  79.82 &  67.96 &  17.26 &  20.01 \\
BERT (large, cased), Layer 3 &  79.50 &  68.82 &  17.42 &  21.83 \\
BERT (large, cased), Layer 4 &  87.49 &  71.13 &  24.06 &  23.21 \\
BERT (large, cased), Layer 5 &  89.81 &  72.06 &  30.27 &  24.13 \\
BERT (large, cased), Layer 6 &  89.92 &  74.30 &  31.44 &  26.75 \\
BERT (large, cased), Layer 7 &  90.39 &  75.93 &  33.27 &  27.74 \\
BERT (large, cased), Layer 8 &  90.28 &  76.99 &  33.34 &  29.94 \\
BERT (large, cased), Layer 9 &  90.09 &  78.87 &  33.16 &  30.07 \\
BERT (large, cased), Layer 10 &  89.92 &  80.08 &  33.31 &  30.17 \\
BERT (large, cased), Layer 11 &  90.20 &  81.23 &  34.49 &  31.78 \\
BERT (large, cased), Layer 12 &  91.22 &  83.00 &  37.27 &  34.10 \\
BERT (large, cased), Layer 13 &  93.04 &  83.66 &  40.10 &  35.04 \\
BERT (large, cased), Layer 14 &  93.64 &  84.11 &  43.11 &  39.67 \\
BERT (large, cased), Layer 15 &  93.18 &  84.21 &  44.92 &  43.12 \\
BERT (large, cased), Layer 16 &  93.14 &  84.34 &  45.37 &  46.54 \\
BERT (large, cased), Layer 17 &  92.80 &  84.44 &  45.60 &  47.76 \\
BERT (large, cased), Layer 18 &  91.72 &  84.03 &  45.82 &  47.34 \\
BERT (large, cased), Layer 19 &  91.48 &  84.29 &  46.46 &  46.00 \\
BERT (large, cased), Layer 20 &  90.78 &  84.25 &  46.07 &  44.81 \\
BERT (large, cased), Layer 21 &  87.97 &  82.36 &  44.53 &  41.91 \\
BERT (large, cased), Layer 22 &  85.19 &  77.58 &  43.03 &  37.49 \\
BERT (large, cased), Layer 23 &  84.23 &  77.02 &  42.00 &  35.21 \\
BERT (large, cased), Layer 24 &  83.30 &  74.83 &  41.29 &  34.38 \\
BERT (large, cased), Scalar Mix &  93.59 &  84.98 &  47.32 &  45.94 \\
\bottomrule
\end{tabular}
\caption{Segmentation task performance of a linear probing model trained on top of the BERT contextualizers.}
\end{table*}

\newpage

\subsection{Pairwise Relations (ELMo and OpenAI Transformer)}

\begin{table*}[h]
\centering
\setlength\tabcolsep{4pt}
\footnotesize
\begin{tabular}{lccccccc}
\toprule
\multirow{2}{*}{Pretrained Representation} & \multicolumn{2}{c}{\begin{tabular}{@{}l@{}}Syntactic Dep. \\ Arc Prediction \end{tabular}} & \multicolumn{2}{c}{\begin{tabular}{@{}l@{}}Syntactic Dep. \\ Arc Classification \end{tabular}} & \multirow{2}{*}{\begin{tabular}{@{}l@{}}Semantic Dep. \\ Arc Prediction \end{tabular}} & \multirow{2}{*}{\begin{tabular}{@{}l@{}}Semantic Dep. \\ Arc Classification \end{tabular}} & \multirow{2}{*}{\begin{tabular}{@{}l@{}} Coreference \\ Arc Prediction \end{tabular}} \\
\cmidrule(lr){2-3} \cmidrule(lr){4-5}
& {PTB} & {EWT} & {PTB} & {EWT} & & &  \\
\midrule
ELMo (original), Layer 0 & 78.27 & 77.73 & 82.05 & 78.52 & 70.65 & 77.48 & 72.89 \\
ELMo (original), Layer 1 & 89.04 & 86.46 & 96.13 & 93.01 & 87.71 & 93.31 & 71.33 \\
ELMo (original), Layer 2 & 88.33 & 85.34 & 94.72 & 91.32 & 86.44 & 90.22 & 68.46 \\
ELMo (original), Scalar Mix & 89.30 & 86.56 & 95.81 & 91.69 & 87.79 & 93.13 & 73.24 \\
\midrule
ELMo (4-layer), Layer 0 & 78.09 & 77.57 & 82.13 & 77.99 & 69.96 & 77.22 & 73.57 \\
ELMo (4-layer), Layer 1 & 88.79 & 86.31 & 96.20 & 93.20 & 87.15 & 93.27 & 72.93 \\
ELMo (4-layer), Layer 2  & 87.33 & 84.75 & 95.38 & 91.87 & 85.29 & 90.57 & 71.78 \\
ELMo (4-layer), Layer 3 & 86.74 & 84.17 & 95.06 & 91.55 & 84.44 & 90.04 & 70.11 \\
ELMo (4-layer), Layer 4 & 87.61 & 85.09 & 94.14 & 90.68 & 85.81 & 89.45 & 68.36 \\
ELMo (4-layer), Scalar Mix & 88.98 & 85.94 & 95.82 & 91.77 & 87.39 & 93.25 & 73.88 \\
\midrule
ELMo (transformer), Layer 0 & 78.10 & 78.04 & 81.09 & 77.67 & 70.11 & 77.11 & 72.50 \\
ELMo (transformer), Layer 1 & 88.24 & 85.48 & 93.62 & 89.18 & 85.16 & 90.66 & 72.47 \\
ELMo (transformer), Layer 2 & 88.87 & 84.72 & 94.14 & 89.40 & 85.97 & 91.29 & 73.03 \\
ELMo (transformer), Layer 3 & 89.01 & 84.62 & 94.07 & 89.17 & 86.83 & 90.35 & 72.62 \\
ELMo (transformer), Layer 4 & 88.55 & 85.62 & 94.14 & 89.00 & 86.00 & 89.04 & 71.80 \\
ELMo (transformer), Layer 5 & 88.09 & 83.23 & 92.70 & 88.84 & 85.79 & 89.66 & 71.62 \\
ELMo (transformer), Layer 6 & 87.22 & 83.28 & 92.55 & 87.13 & 84.71 & 87.21 & 66.35 \\
ELMo (transformer), Scalar Mix & 90.74 & 86.39 & 96.40 & 91.06 & 89.18 & 94.35 & 75.52 \\
\midrule
OpenAI transformer, Layer 0 & 80.80 & 79.10 & 83.35 & 80.32 & 76.39 & 80.50 & 72.58 \\
OpenAI transformer, Layer 1 & 81.91 & 79.99 & 88.22 & 84.51 & 77.70 & 83.88 & 75.23 \\
OpenAI transformer, Layer 2 & 82.56 & 80.22 & 89.34 & 85.99 & 78.47 & 85.85 & 75.77 \\
OpenAI transformer, Layer 3 & 82.87 & 81.21 & 90.89 & 87.67 & 78.91 & 87.76 & 75.81 \\
OpenAI transformer, Layer 4 & 83.69 & 82.07 & 92.21 & 89.24 & 80.51 & 89.59 & 75.99 \\
OpenAI transformer, Layer 5 & 84.53 & 82.77 & 93.12 & 90.34 & 81.95 & 90.25 & 76.05 \\
OpenAI transformer, Layer 6 & 85.47 & 83.89 & 93.71 & 90.63 & 83.88 & 90.99 & 74.43 \\
OpenAI transformer, Layer 7 & 86.32 & 84.15 & 93.95 & 90.82 & 85.15 & 91.18 & 74.05 \\
OpenAI transformer, Layer 8 & 86.84 & 84.06 & 94.16 & 91.02 & 85.23 & 90.86 & 74.20 \\
OpenAI transformer, Layer 9 & 87.00 & 84.47 & 93.95 & 90.77 & 85.95 & 90.85 & 74.57 \\
OpenAI transformer, Layer 10 & 86.76 & 84.28 & 93.40 & 90.26 & 85.17 & 89.94 & 73.86 \\
OpenAI transformer, Layer 11 & 85.84 & 83.42 & 92.82 & 89.07 & 83.39 & 88.46 & 72.03 \\
OpenAI transformer, Layer 12 & 85.06 & 83.02 & 92.37 & 89.08 & 81.88 & 87.47 & 70.44 \\
OpenAI transformer, Scalar Mix & 87.18 & 85.30 & 94.51 & 91.55 & 86.13 & 91.55 & 76.47 \\
\midrule
GloVe (840B.300d) & 74.14 & 73.94 & 77.54 & 72.74 & 68.94 & 71.84 & 72.96 \\
\bottomrule
\end{tabular}
\caption{Pairwise relation task performance of a linear probing model trained on top of the ELMo and OpenAI contextualizers, compared against a GloVe-based probing baseline.}
\end{table*}

\newpage

\subsection{Pairwise Relations (BERT)}

\begin{table*}[h]
\centering
\setlength\tabcolsep{4pt}
\footnotesize
\begin{tabular}{lccccccc}
\toprule
\multirow{2}{*}{Pretrained Representation} & \multicolumn{2}{c}{\begin{tabular}{@{}l@{}}Syntactic Dep. \\ Arc Prediction \end{tabular}} & \multicolumn{2}{c}{\begin{tabular}{@{}l@{}}Syntactic Dep. \\ Arc Classification \end{tabular}} & \multirow{2}{*}{\begin{tabular}{@{}l@{}}Semantic Dep. \\ Arc Prediction \end{tabular}} & \multirow{2}{*}{\begin{tabular}{@{}l@{}}Semantic Dep. \\ Arc Classification \end{tabular}} & \multirow{2}{*}{\begin{tabular}{@{}l@{}} Coreference \\ Arc Prediction \end{tabular}} \\
\cmidrule(lr){2-3} \cmidrule(lr){4-5}
& {PTB} & {EWT} & {PTB} & {EWT} & & &  \\
\midrule
BERT (base, cased), Layer 0 & 83.00 & 80.36 & 83.47 & 79.15 & 80.26 & 80.35 & 74.93 \\
BERT (base, cased), Layer 1 & 83.66 & 81.69 & 86.92 & 82.62 & 80.81 & 82.69 & 75.35 \\
BERT (base, cased), Layer 2 & 84.00 & 82.66 & 91.90 & 88.51 & 79.34 & 87.45 & 75.19 \\
BERT (base, cased), Layer 3 & 84.12 & 82.86 & 92.80 & 89.49 & 79.05 & 88.41 & 75.83 \\
BERT (base, cased), Layer 4 & 85.50 & 84.07 & 93.91 & 91.02 & 81.37 & 90.20 & 76.14 \\
BERT (base, cased), Layer 5 & 86.67 & 84.69 & 94.87 & 92.01 & 83.41 & 91.34 & 76.35 \\
BERT (base, cased), Layer 6 & 87.98 & 85.91 & 95.57 & 93.01 & 85.73 & 92.47 & 75.95 \\
BERT (base, cased), Layer 7 & 88.24 & 86.30 & 95.65 & 93.31 & 85.96 & 92.75 & 75.37 \\
BERT (base, cased), Layer 8 & 88.64 & 86.49 & 95.90 & 93.39 & 86.59 & 93.18 & 76.39 \\
BERT (base, cased), Layer 9 & 88.76 & 86.17 & 95.84 & 93.32 & 86.74 & 92.68 & 76.62 \\
BERT (base, cased), Layer 10 & 88.16 & 85.86 & 95.42 & 92.82 & 86.29 & 91.79 & 76.84 \\
BERT (base, cased), Layer 11 & 87.74 & 85.40 & 95.09 & 92.37 & 85.83 & 91.07 & 76.88 \\
BERT (base, cased), Layer 12 & 85.93 & 83.99 & 94.79 & 91.70 & 82.71 & 90.10 & 76.78 \\
BERT (base, cased), Scalar Mix & 89.06 & 86.58 & 95.91 & 93.10 & 87.10 & 93.38 & 77.88 \\
\midrule
BERT (large, cased), Layer 0 & 82.22 & 79.92 & 83.57 & 79.32 & 79.04 & 81.25 & 73.75 \\
BERT (large, cased), Layer 1 & 81.65 & 80.04 & 85.23 & 80.95 & 77.97 & 81.36 & 73.99 \\
BERT (large, cased), Layer 2 & 81.84 & 80.09 & 87.39 & 83.80 & 77.17 & 82.44 & 73.89 \\
BERT (large, cased), Layer 3 & 81.66 & 80.35 & 87.36 & 83.74 & 76.92 & 82.91 & 73.62 \\
BERT (large, cased), Layer 4 & 83.56 & 82.17 & 91.44 & 88.45 & 78.43 & 87.32 & 72.99 \\
BERT (large, cased), Layer 5 & 84.24 & 82.94 & 92.33 & 89.62 & 79.28 & 88.85 & 73.34 \\
BERT (large, cased), Layer 6 & 85.05 & 83.50 & 93.75 & 91.02 & 80.18 & 90.14 & 74.02 \\
BERT (large, cased), Layer 7 & 85.43 & 84.03 & 94.06 & 91.65 & 80.64 & 90.69 & 74.55 \\
BERT (large, cased), Layer 8 & 85.41 & 83.92 & 94.18 & 91.66 & 80.64 & 90.82 & 75.92 \\
BERT (large, cased), Layer 9 & 85.35 & 83.76 & 94.11 & 91.10 & 80.64 & 90.62 & 76.00 \\
BERT (large, cased), Layer 10 & 85.51 & 83.92 & 94.09 & 91.17 & 81.51 & 90.43 & 76.19 \\
BERT (large, cased), Layer 11 & 85.91 & 83.88 & 94.48 & 91.73 & 82.05 & 91.13 & 75.86 \\
BERT (large, cased), Layer 12 & 86.80 & 85.13 & 95.03 & 92.37 & 83.99 & 92.08 & 75.13 \\
BERT (large, cased), Layer 13 & 87.64 & 86.00 & 95.54 & 93.02 & 84.91 & 92.74 & 74.63 \\
BERT (large, cased), Layer 14 & 88.62 & 86.50 & 95.94 & 93.62 & 85.91 & 93.51 & 75.16 \\
BERT (large, cased), Layer 15 & 88.87 & 86.95 & 96.02 & 93.66 & 86.49 & 93.86 & 75.58 \\
BERT (large, cased), Layer 16 & 89.36 & 87.25 & 96.18 & 93.86 & 87.79 & 93.83 & 75.15 \\
BERT (large, cased), Layer 17 & 89.62 & 87.47 & 96.01 & 93.88 & 88.14 & 93.41 & 75.93 \\
BERT (large, cased), Layer 18 & 89.41 & 87.00 & 95.82 & 93.47 & 87.77 & 93.00 & 77.85 \\
BERT (large, cased), Layer 19 & 88.78 & 86.60 & 95.59 & 92.98 & 87.16 & 92.27 & 80.47 \\
BERT (large, cased), Layer 20 & 88.24 & 85.87 & 95.12 & 92.47 & 86.45 & 91.33 & 80.94 \\
BERT (large, cased), Layer 21 & 86.48 & 84.21 & 94.21 & 91.12 & 83.94 & 89.42 & 81.14 \\
BERT (large, cased), Layer 22 & 85.42 & 83.24 & 92.94 & 90.02 & 82.01 & 88.17 & 80.36 \\
BERT (large, cased), Layer 23 & 84.69 & 82.81 & 92.28 & 89.47 & 81.07 & 87.32 & 79.64 \\
BERT (large, cased), Layer 24 & 83.24 & 81.48 & 91.07 & 87.88 & 78.24 & 85.98 & 79.35 \\
BERT (large, cased), Scalar Mix & 90.09 & 87.51 & 96.15 & 93.61 & 88.49 & 94.25 & 81.16 \\
\bottomrule
\end{tabular}
\caption{Pairwise relation task performance of a linear probing model trained on top of the BERT contextualizers.}
\end{table*}

\newpage

\labeledsection{Full Results for Transferring Between Pretraining Tasks}{full_results_transferring_pretraining_tasks}

\subsection{Token Labeling}

\begin{table*}[!htbp]
\centering
\setlength\tabcolsep{4pt}
\resizebox*{!}{0.88\textheight}{%
\begin{tabular}{lccccc}
\toprule
\multirow{2}{*}{Pretrained Representation} & & & \multicolumn{2}{c}{Supersense ID} & \\
\cmidrule{4-5}
& POS (EWT) & {ST} & {PS-Role} & {PS-Fxn} & {EF} \\
\midrule
Untrained ELMo (original), Layer 0 & 77.05 & 76.09 & 36.99 & 48.17 & 43.08 \\
Untrained ELMo (original), Layer 1 & 56.03 & 68.63 & 16.01 & 24.71 & 45.57 \\
Untrained ELMo (original), Layer 2 & 55.89 & 68.51 & 16.01 & 25.44 & 46.06 \\
Untrained ELMo (original), Scalar Mix & 78.58 & 82.45 & 38.23 & 48.90 & 47.37 \\
\midrule
CCG, Layer 0 & 84.33 & 79.53 & 38.38 & 53.29 & 47.71 \\
CCG, Layer 1 & 88.02 & 87.97 & 46.27 & 58.48 & 57.96 \\
CCG, Layer 2 & 87.81 & 87.38 & 43.79 & 58.55 & 57.98 \\
CCG, Scalar Mix & 90.44 & 91.21 & 50.07 & 65.57 & 60.24 \\
\midrule
Chunk, Layer 0 & 82.51 & 78.45 & 37.06 & 49.12 & 38.93 \\
Chunk, Layer 1 & 87.33 & 87.42 & 44.81 & 59.36 & 55.66 \\
Chunk, Layer 2 & 86.61 & 87.04 & 39.91 & 58.11 & 56.95 \\
Chunk, Scalar Mix & 88.62 & 89.77 & 44.23 & 60.01 & 56.24 \\
\midrule
PTB (POS), Layer 0 & 84.58 & 79.95 & 37.43 & 49.49 & 46.19 \\
PTB (POS), Layer 1 & 90.53 & 90.10 & 42.47 & 59.80 & 61.28 \\
PTB (POS), Layer 2 & 90.45 & 89.83 & 44.37 & 58.92 & 62.14 \\
PTB (POS), Scalar Mix & 90.75 & 91.13 & 45.39 & 60.67 & 62.77 \\
\midrule
Parent, Layer 0 & 81.84 & 78.47 & 36.33 & 49.71 & 38.35 \\
Parent, Layer 1 & 87.21 & 87.36 & 45.98 & 58.85 & 54.45 \\
Parent, Layer 2 & 86.57 & 86.18 & 42.69 & 58.48 & 54.58 \\
Parent, Scalar Mix & 89.10 & 90.01 & 44.88 & 61.92 & 55.64 \\
\midrule
GParent, Layer 0 & 81.85 & 78.77 & 37.06 & 51.75 & 40.46 \\
GParent, Layer 1 & 86.05 & 86.78 & 46.86 & 60.82 & 55.58 \\
GParent, Layer 2 & 85.64 & 86.17 & 45.25 & 62.13 & 55.65 \\
GParent, Scalar Mix & 88.08 & 89.48 & 48.03 & 63.38 & 55.96 \\
\midrule
GGParent, Layer 0 & 81.44 & 77.88 & 38.74 & 49.12 & 42.17 \\
GGParent, Layer 1 & 83.51 & 85.23 & 44.08 & 57.68 & 55.77 \\
GGParent, Layer 2 & 83.17 & 84.10 & 39.40 & 56.29 & 55.82 \\
GGParent, Scalar Mix & 86.18 & 88.84 & 44.52 & 61.62 & 55.50 \\
\midrule
Syn.~Arc Prediction (PTB), Layer 0 & 79.97 & 77.34 & 36.26 & 47.15 & 38.81 \\
Syn.~Arc Prediction (PTB), Layer 1 & 80.67 & 82.60 & 40.06 & 54.61 & 47.86 \\
Syn.~Arc Prediction (PTB), Layer 2 & 78.83 & 80.91 & 34.65 & 52.12 & 45.64
\\
Syn.~Arc Prediction (PTB), Scalar Mix & 85.76 & 88.13 & 40.79 & 54.17 & 50.91 \\
\midrule
Syn.~Arc Classification (PTB), Layer 0 & 83.61 & 79.61 & 37.21 & 51.97 & 42.07 \\
Syn.~Arc Classification (PTB), Layer 1 & 89.28 & 88.70 & 47.22 & 61.11 & 55.55 \\
Syn.~Arc Classification (PTB), Layer 2 & 88.77 & 88.12 & 44.66 & 58.92 & 56.16 \\
Syn.~Arc Classification (PTB), Scalar Mix & 90.18 & 90.99 & 48.17 & 62.21 & 56.90 \\
\midrule
Sem.~Arc Prediction, Layer 0 & 78.64 & 76.95 & 34.43 & 49.78 & 39.64 \\
Sem.~Arc Prediction, Layer 1 & 74.66 & 74.83 & 33.92 & 47.88 & 36.46 \\
Sem.~Arc Prediction, Layer 2 & 74.06 & 73.42 & 30.85 & 45.39 & 35.63 \\
Sem.~Arc Prediction, Scalar Mix & 83.77 & 85.06 & 38.45 & 57.16 & 48.27 \\
\midrule
Sem.~Arc Classification, Layer 0 & 83.17 & 79.17 & 38.60 & 51.54 & 44.79 \\
Sem.~Arc Classification, Layer 1 & 86.45 & 87.04 & 44.81 & 58.19 & 55.18 \\
Sem.~Arc Classification, Layer 2 & 85.42 & 85.87 & 41.45 & 58.55 & 52.87 \\
Sem.~Arc Classification, Scalar Mix & 88.44 & 90.00 & 45.03 & 61.33 & 56.07 \\
\midrule
Conj, Layer 0 & 72.21 & 73.87 & 37.43 & 47.95 & 36.33 \\
Conj, Layer 1 & 64.95 & 68.96 & 27.70 & 41.89 & 42.10 \\
Conj, Layer 2 & 64.03 & 67.17 & 27.56 & 37.21 & 40.59 \\
Conj, Scalar Mix & 76.96 & 80.22 & 36.33 & 50.66 & 42.79 \\
\midrule
BiLM, Layer 0 & 87.54 & 90.22 & 50.88 & 67.32 & 59.65 \\
BiLM, Layer 1 & 86.55 & 87.19 & 50.22 & 67.11 & 59.32 \\
BiLM, Layer 2 & 86.49 & 89.67 & 49.34 & 66.01 & 59.45 \\
BiLM, Scalar Mix & 86.76 & 90.11 & 50.44 & 67.32 & 67.32 \\
\midrule
ELMo (original), Layer 0 & 89.71 & 83.99 & 41.45 & 52.41 & 52.49 \\
ELMo (original), Layer 1 & 95.61 & 93.82 & 74.12 & 84.87 & 73.20 \\
ELMo (original), Layer 2 & 94.52 & 92.41 & 75.44 & 83.11 & 72.11 \\
ELMo (original), Scalar Mix & 95.09 & 93.86 & 74.56 & 84.65 & 84.65 \\
\midrule
GloVe (840B.300d) & 83.93 & 80.92 & 40.79 & 51.54 & 49.70 \\
\bottomrule
\end{tabular}}
\caption{Target token labeling task performance of contextualizers pretrained on a variety of different tasks. The probing model used is linear, and the contextualizer architecture is ELMo (original).}
\end{table*}

\newpage

\subsection{Segmentation}

\begin{table*}[!htp]
\centering
\setlength\tabcolsep{4pt}
\resizebox*{!}{0.91\textheight}{%
\begin{tabular}{lcc}
\toprule
{Pretrained Representation} & NER & GED \\
\midrule
Untrained ELMo (original), Layer 0 & 24.71 & 0.00 \\
Untrained ELMo (original), Layer 1 & 0.00 & 0.00 \\
Untrained ELMo (original), Layer 2 & 0.00 & 0.00 \\
Untrained ELMo (original), Scalar Mix & 34.28 & 1.81\\
\midrule
CCG, Layer 0 & 32.30 & 8.89 \\
CCG, Layer 1 & 44.01 & 22.68 \\
CCG, Layer 2 & 42.45 & 25.15 \\
CCG, Scalar Mix & 49.07 & 4.52 \\
\midrule
Chunk, Layer 0 & 23.47 & 5.80 \\
Chunk, Layer 1 & 45.44 & 5.46 \\
Chunk, Layer 2 & 43.59 & 24.11 \\
Chunk, Scalar Mix & 46.83 & 4.30 \\
\midrule
PTB (POS), Layer 0 & 32.64 & 7.87 \\
PTB (POS), Layer 1 & 52.03 & 5.80 \\
PTB (POS), Layer 2 & 52.04 & 9.76 \\
PTB (POS), Scalar Mix & 53.51 & 3.19 \\
\midrule
Parent, Layer 0 & 25.11 & 6.66 \\
Parent, Layer 1 & 42.76 & 6.22 \\
Parent, Layer 2 & 42.49 & 8.33 \\
Parent, Scalar Mix & 47.06 & 3.01 \\
\midrule
GParent, Layer 0 & 30.39 & 4.58 \\
GParent, Layer 1 & 47.67 & 6.20 \\
GParent, Layer 2 & 47.87 & 10.34 \\
GParent, Scalar Mix & 50.06 & 1.71 \\
\midrule
GGParent, Layer 0 & 28.57 & 2.25 \\
GGParent, Layer 1 & 46.21 & 4.32 \\
GGParent, Layer 2 & 45.34 & 3.74 \\
GGParent, Scalar Mix & 48.19 & 1.54 \\
\midrule
Syn.~Arc Prediction (PTB), Layer 0 & 26.77 & 1.82 \\
Syn.~Arc Prediction (PTB), Layer 1 & 43.93 & 5.94 \\
Syn.~Arc Prediction (PTB), Layer 2 & 41.83 & 14.50 \\
Syn.~Arc Prediction (PTB), Scalar Mix & 46.58 & 1.47 \\
\midrule
Syn.~Arc Classification (PTB), Layer 0 & 33.10 & 3.51 \\
Syn.~Arc Classification (PTB), Layer 1 & 50.76 & 3.92 \\
Syn.~Arc Classification (PTB), Layer 2 & 49.64 & 5.77 \\
Syn.~Arc Classification (PTB), Scalar Mix & 53.00 & 1.27 \\
\midrule
Sem.~Arc Prediction, Layer 0 & 24.47 & 1.05 \\
Sem.~Arc Prediction, Layer 1 & 34.47 & 10.78 \\
Sem.~Arc Prediction, Layer 2 & 31.30 & 10.77 \\
Sem.~Arc Prediction, Scalar Mix & 36.97 & 0.32 \\
\midrule
Sem.~Arc Classification, Layer 0 & 34.00 & 5.08 \\
Sem.~Arc Classification, Layer 1 & 48.07 & 5.39 \\
Sem.~Arc Classification, Layer 2 & 46.67 & 6.24 \\
Sem.~Arc Classification, Scalar Mix & 50.80 & 1.75 \\
\midrule
Conj, Layer 0 & 17.15 & 3.99 \\
Conj, Layer 1 & 37.61 & 0.87 \\
Conj, Layer 2 & 34.78 & 2.38 \\
Conj, Scalar Mix & 40.97 & 0.33 \\
\midrule
BiLM, Layer 0 & 56.05 & 3.99 \\
BiLM, Layer 1 & 57.19 & 1.22 \\
BiLM, Layer 2 & 57.05 & 1.03 \\
BiLM, Scalar Mix & 58.50 & 1.29 \\
\midrule
ELMo (original), Layer 0 & 64.39 & 18.49 \\
ELMo (original), Layer 1 & 82.85 & 29.37 \\
ELMo (original), Layer 2 & 82.80 & 26.08 \\
ELMo (original), Scalar Mix & 82.90 & 27.54 \\
\midrule
GloVe (840B.300d) & 53.22 & 14.94 \\
\bottomrule
\end{tabular}}
\caption{Target segmentation task performance of contextualizers pretrained on a variety of different tasks. The probing model used is linear, and the contextualizer architecture is ELMo (original).}
\end{table*}

\newpage

\subsection{Pairwise Prediction}

\begin{table*}[!htp]
\centering
\resizebox*{!}{0.91\textheight}{%
\setlength\tabcolsep{4pt}
\footnotesize
\begin{tabular}{lccc}
\toprule
\multirow{4}{*}{Pretrained Representation} & & & \\
& {\begin{tabular}{@{}c@{}}Syn.~Arc \\ Prediction \\ (EWT)\end{tabular}}  & {\begin{tabular}{@{}c@{}}Syn.~Arc \\ Classification \\ (EWT)\end{tabular}}  & {{\begin{tabular}{@{}c@{}}Coreference \\ Arc Prediction\end{tabular}}} \\
\midrule
Untrained ELMo (original), Layer 0 & 73.75 & 66.27 & 66.25 \\
Untrained ELMo (original), Layer 1 & 68.40 & 56.73 & 62.82 \\
Untrained ELMo (original), Layer 2 & 68.86 & 56.62 & 63.15 \\
Untrained ELMo (original), Scalar Mix & 72.24 & 70.62 & 69.72 \\
\midrule
CCG, Layer 0 & 75.92 & 69.84 & 67.84 \\
CCG, Layer 1 & 84.93 & 85.59 & 62.10 \\
CCG, Layer 2 & 84.45 & 84.59 & 59.19 \\
CCG, Scalar Mix & 85.44 & 88.11 & 70.14 \\
\midrule
Chunk, Layer 0 & 76.67 & 69.72 & 65.60 \\
Chunk, Layer 1 & 85.18 & 86.50 & 62.74 \\
Chunk, Layer 2 & 84.80 & 84.84 & 60.23 \\
Chunk, Scalar Mix & 85.42 & 87.57 & 68.92 \\
\midrule
PTB (POS), Layer 0 & 76.07 & 70.32 & 67.50 \\
PTB (POS), Layer 1 & 83.97 & 86.64 & 63.43 \\
PTB (POS), Layer 2 & 83.88 & 86.44 & 61.61 \\
PTB (POS), Scalar Mix & 84.17 & 87.72 & 69.61 \\
\midrule
Parent, Layer 0 & 76.20 & 68.99 & 67.80 \\
Parent, Layer 1 & 84.93 & 86.15 & 62.69 \\
Parent, Layer 2 & 85.57 & 85.61 & 59.10 \\
Parent, Scalar Mix & 86.01 & 87.49 & 69.34 \\
\midrule
GParent, Layer 0 & 76.59 & 69.51 & 68.99 \\
GParent, Layer 1 & 85.96 & 85.33 & 60.84 \\
GParent, Layer 2 & 85.69 & 84.38 & 58.76 \\
GParent, Scalar Mix & 86.17 & 87.49 & 70.24 \\
\midrule
GGParent, Layer 0 & 76.28 & 69.91 & 69.24 \\
GGParent, Layer 1 & 85.74 & 83.45 & 59.73 \\
GGParent, Layer 2 & 85.49 & 82.12 & 58.89 \\
GGParent, Scalar Mix & 86.27 & 86.57 & 70.58 \\
\midrule
Syn.~Arc Prediction (PTB), Layer 0 & 77.04 & 68.01 & 68.28 \\
Syn.~Arc Prediction (PTB), Layer 1 & 90.39 & 81.00 & 60.29 \\
Syn.~Arc Prediction (PTB), Layer 2 & 90.82 & 76.50 & 57.46 \\
Syn.~Arc Prediction (PTB), Scalar Mix & 91.66 & 84.18 & 69.15 \\
\midrule
Syn.~Arc Classification (PTB), Layer 0 & 76.14 & 71.80 & 68.40 \\
Syn.~Arc Classification (PTB), Layer 1 & 86.55 & 90.04 & 62.10 \\
Syn.~Arc Classification (PTB), Layer 2 & 87.46 & 89.35 & 59.74 \\
Syn.~Arc Classification (PTB), Scalar Mix & 87.78 & 90.98 & 70.00 \\
\midrule
Sem.~Arc Prediction, Layer 0 & 76.25 & 67.73 & 69.44 \\
Sem.~Arc Prediction, Layer 1 & 84.91 & 73.11 & 57.62 \\
Sem.~Arc Prediction, Layer 2 & 85.86 & 69.75 & 55.91 \\
Sem.~Arc Prediction, Scalar Mix & 86.37 & 80.74 & 69.72 \\
\midrule
Sem.~Arc Classification, Layer 0 & 75.85 & 70.12 & 68.96 \\
Sem.~Arc Classification, Layer 1 & 85.30 & 86.21 & 60.25 \\
Sem.~Arc Classification, Layer 2 & 86.10 & 84.50 & 58.39 \\
Sem.~Arc Classification, Scalar Mix & 86.53 & 87.75 & 70.36 \\
\midrule
Conj, Layer 0 & 72.62 & 58.40 & 68.50 \\
Conj, Layer 1 & 80.84 & 68.12 & 58.46 \\
Conj, Layer 2 & 80.46 & 64.30 & 57.89 \\
Conj, Scalar Mix & 80.96 & 73.89 & 71.96 \\
\midrule
BiLM, Layer 0 & 84.27 & 86.74 & 71.75 \\
BiLM, Layer 1 & 86.36 & 86.86 & 70.47 \\
BiLM, Layer 2 & 86.44 & 86.19 & 70.14 \\
BiLM, Scalar Mix & 86.42 & 85.93 & 71.62 \\
\midrule
ELMo (original), Layer 0 & 77.73 & 78.52 & 72.89 \\
ELMo (original), Layer 1 & 86.46 & 93.01 & 71.33 \\
ELMo (original), Layer 2 & 85.34 & 91.32 & 68.46 \\
ELMo (original), Scalar Mix & 86.56 & 91.69 & 73.24 \\
\midrule
GloVe (840B.300d) & 73.94 & 72.74 & 72.96 \\
\bottomrule
\end{tabular}}
\caption{Target pairwise prediction task performance of contextualizers pretrained on a variety of different tasks. The probing model used is linear, and the contextualizer architecture is ELMo (original).}
\end{table*}

\end{appendices}
\end{document}